\documentclass[letterpaper]{article} 
\usepackage[submission]{aaai23}  
\usepackage{times}  
\usepackage{helvet}  
\usepackage{courier}  
\usepackage[hyphens]{url}  
\usepackage{graphicx} 
\usepackage{natbib}  
\usepackage{caption} 
\usepackage{algorithm}
\usepackage{algorithmic}

\usepackage{mathrsfs}
\usepackage{amssymb}
\usepackage{bm}
\usepackage{amsmath}
\usepackage{dsfont}
\usepackage{booktabs}
\usepackage{epstopdf}
\usepackage{multirow}
\usepackage{comment}
\usepackage[table,xcdraw]{xcolor}
\usepackage{cite}
\usepackage{endnotes}
\usepackage{makecell}

\usepackage{newfloat}
\usepackage{listings}
\DeclareCaptionStyle{ruled}{labelfont=normalfont,labelsep=colon,strut=off} 
\floatstyle{ruled}
\newfloat{listing}{tb}{lst}{}
\floatname{listing}{Listing}

\title{Hard Sample Aware Network for Contrastive Deep Graph Clustering\footnotemark[5]}
\author{Yue Liu,\textsuperscript{\rm 1}\footnotemark[1] Xihong Yang,\textsuperscript{\rm 1}\footnotemark[1], Sihang Zhou,\textsuperscript{\rm 2} Xinwang Liu,\textsuperscript{\rm 1} \footnotemark[2] Zhen Wang,\textsuperscript{\rm 3} \\Ke Liang,\textbf{\textsuperscript{\rm 1}} Wenxuan Tu,\textsuperscript{\rm 1} Liang Li,\textsuperscript{\rm 1} Jingcan Duan,\textsuperscript{\rm 1} Cancan Chen\textsuperscript{\rm 4}}
\affiliations{
    \textsuperscript{\rm 1}College of Computer, National University of Defense Technology \\
    \textsuperscript{\rm 2}College of Intelligence Science and Technology, National University of Defense Technology\\
    \textsuperscript{\rm 3}Northwestern Polytechnical University \ 
    \textsuperscript{\rm 4}Beijing Information Science and Technology University  

    yueliu@nudt.edu.cn

}

\usepackage{bibentry}

\begin{document}

\maketitle

\begin{abstract}

\footnotetext[1]{Equal contribution}
\footnotetext[2]{Corresponding author}
\footnotetext[5]{Accepted by AAAI 2023. Pre-print version.}

Contrastive deep graph clustering, which aims to divide nodes into disjoint groups via contrastive mechanisms, is a challenging research spot. Among the recent works, hard sample mining-based algorithms have achieved great attention for their promising performance. However, we find that the existing hard sample mining methods have two problems as follows. 1) In the hardness measurement, the important structural information is overlooked for similarity calculation, degrading the representativeness of the selected hard negative samples. 2) Previous works merely focus on the hard negative sample pairs while neglecting the hard positive sample pairs. Nevertheless, samples within the same cluster but with low similarity should also be carefully learned. To solve the problems, we propose a novel contrastive deep graph clustering method dubbed Hard Sample Aware Network (HSAN) by introducing a comprehensive similarity measure criterion and a general dynamic sample weighing strategy. Concretely, in our algorithm, the similarities between samples are calculated by considering both the attribute embeddings and the structure embeddings, better revealing sample relationships and assisting hardness measurement. Moreover, under the guidance of the carefully collected high-confidence clustering information, our proposed weight modulating function will first recognize the positive and negative samples and then dynamically up-weight the hard sample pairs while down-weighting the easy ones. In this way, our method can mine not only the hard negative samples but also the hard positive sample, thus improving the discriminative capability of the samples further. Extensive experiments and analyses demonstrate the superiority and effectiveness of our proposed method. The source code of HSAN is shared at https://github.com/yueliu1999/HSAN and a collection (papers, codes and, datasets) of deep graph clustering is shared at https://github.com/yueliu1999/Awesome-Deep-Graph-Clustering on Github.

\end{abstract}

\section{Introduction}
In recent years, contrastive learning has achieved promising performance in the field of deep graph clustering benefiting from the powerful potential supervision information extraction capability \cite{DCRN,AGC-DRR}. Among the recent works, researchers demonstrate the effectiveness of hard negative sample mining in contrastive learning. Concretely, GDCL \cite{GDCL} is proposed to correct the bias of the negative sample selection via clustering pseudo labels. ProGCL \cite{ProGCL} first filters the false negative samples and then generates a more abundant negative sample set by the sample interpolation.



Although verified to be effective, we point out two drawbacks in the existing methods as follows. 1) While measuring the hardness of samples, the important structural information is neglected for the sample similarity calculation, degrading the representativeness of the selected hard negative samples. 2) Previous works only focus on the hard negative samples while overlooking the hard positive samples, limiting the discriminative capability of samples. We argue that the samples within the same cluster but with low similarity should also be carefully learned.




\begin{figure}[h]
\centering
\small
\begin{minipage}{0.48\linewidth}
\centerline{\includegraphics[width=1\textwidth]{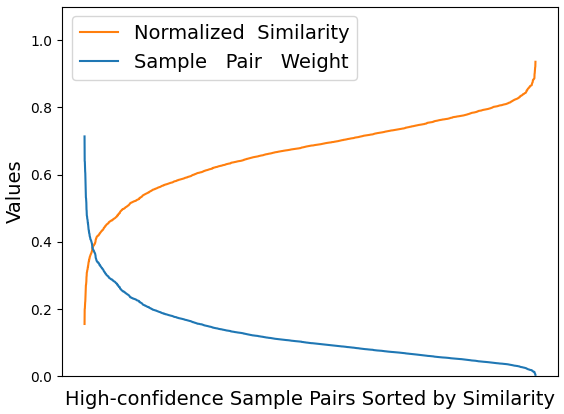}}
\centerline{(a) Positive Sample Pairs}
\vspace{3pt}
\end{minipage}
\begin{minipage}{0.48\linewidth}
\centerline{\includegraphics[width=1\textwidth]{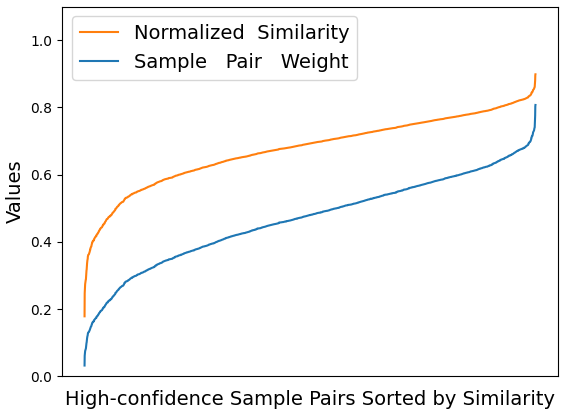}}
\centerline{(b) Negative Sample Pairs}
\vspace{3pt}
\end{minipage}
\caption{Sample weighing strategy illustration. In the sample weighting process, according to the high-confidence pseudo labels generated by the integrated clustering procedure, samples from the same clusters are recognized as potential positive sample pairs. Meanwhile, samples from different clusters are combined as potential negative sample pairs. To drive the network to focus more on the hard samples, we assign larger weights to the positive sample pairs with smaller similarities shown in sub-figure (a) and assign smaller weights to the negative sample pairs with small similarities shown in sub-figure (b). In this way, samples in the same cluster with low similarity and those in different clusters with large similarity are mined as hard samples.}
\label{motivation}
\end{figure}

To solve the mentioned problems, we propose a novel contrastive deep graph clustering method termed Hard Sample Aware Network (HSAN) by designing a comprehensive similarity measure criterion and a general dynamic sample weighting strategy. Concretely, to provide a more reliable hardness measure criterion, the sample similarity is calculated by a learnable linear combination of the attribute similarity and the structural similarity. Besides, a novel contrastive sample weighting strategy is proposed to improve the discriminative capability of the network. Firstly, we perform the clustering algorithm on the consensus node embeddings and generate the high-confidence clustering pseudo labels. Then, samples from the same clusters are recognized as potential positive sample pairs and those from different clusters are selected as potential negative ones. Especially, an adaptive sample weighing function tunes the weights of high-confidence positive and negative sample pairs according to the training difficulty. As illustrated in  Figure \ref{motivation}, positive sample pairs with small similarity and negative ones with large similarity are the hard samples to which more attention should be paid. The main contributions of this paper are summarized as follows.

\begin{itemize}
\item We propose a novel contrastive deep graph clustering termed Hard Sample Aware Network (HSAN). It guides the network to focus on both hard positive and negative sample pairs.

\item To assist the hard sample mining, we design a comprehensive similarity measure criterion by considering both attribute and structure information. It better reveals the similarity between samples.

\item Under the guidance of the high-confidence clustering information, the proposed sample weight modulating strategy dynamically up-weights hard sample pairs while down-weighting the easy samples, thus improving the discriminative capability of the network.

\item Extensive experimental results on six datasets demonstrate the superiority and the effectiveness of our proposed method.

\end{itemize}

\section{Related Work}
\subsection{Deep Graph Clustering}
Deep learning has been successful in many domain including computer vision \cite{ZHOU_3,wanglikang_1,wanglikang_2,wanglikang_3}, time series analysis \cite{xiefeng_1,liumeng_1,liumeng_2,liumeng_3}, bioinformatics \cite{jun_survey,zhangyang_2,tan_1}, and graph data mining \cite{wangaug1, wangaug2,zeng_1,zeng_2,lirong_1,jingcan_1,SHGP,KGR_survey}. Among these directions, deep graph clustering, which aims to encode nodes with neural networks and divide them into disjoint clusters, has attracted great attention in recent years. According to the learning mechanisms, the existing methods can be roughly categorized into three classes: generative methods \cite{MGAE,R-GAE,xiawei_1}, adversarial methods \cite{ARGA,AGC-DRR}, and contrastive methods. More information about the fast-growing deep graph clustering can be checked in our survey paper \cite{deep_graph_clustering_survey}. In this work, we focus on the last category, i.e., the contrastive deep graph clustering. Recently, contrastive mechanisms have succeeded in many domains such as images \cite{yongjie_1} and graphs \cite{GRACE,lirong_survey}, and knowledge graphs \cite{KGESymCL}. Inspired by their success, the contrastive deep graph clustering methods are increasingly proposed. A pioneer AGE \cite{AGE} conducts contrastive learning by a designed adaptive encoder. MVGRL \cite{MVGRL} adopts the InfoMax loss \cite{DIM} to maximize the cross-view mutual information. Subsequently, SCAGC \cite{SCAGC} pulls together the positive samples while pushing away negative ones across augmented views. After that, DCRN and its variants \cite{DCRN,IDCRN} alleviate the collapsed representation by reducing correlation in a dual manner. Then SCGC \cite{SCGC} is proposed to reduce high time cost of the existing methods by simplifying the data augmentation and the graph convolutional operation. More recently, the selection of positive and negative samples has attracted great attention. Concretely, GDCL \cite{GDCL} develops a debiased sampling strategy to correct the bias for negative samples. However, most of the existing methods treat the easy and hard samples equally, leading to indiscriminative capability. To solve this problem, we propose a novel contrastive deep graph clustering by mining the hard samples. In our proposed method, the weights of the hard positive and negative samples will dynamically increased while the weights of the easy ones will be decreased.





\subsection{Hard Sample Mining}
In the contrastive learning methods, one key factor of promising performance is the positive and negative sample selection. Previous works \cite{DCL,HCL,MoCHi} on images have demonstrated that the hard negative samples are hard yet useful. Motivated by their success, more researchers take attention to the hard negative sample mining on graphs. Concretely, GDCL \cite{GDCL} utilizes the clustering pseudo labels to correct the bias of the negative sample selection in the attribute graph clustering task. Besides, CuCo \cite{CuCo} selects and learns the samples from easy to hard in the graph classification task. In addition, to better mine true and hard negative samples, STENCIL \cite{STENCIL} advocates enriching the model with local structure patterns of heterogeneous graphs. More recently, ProGCL \cite{ProGCL} built a more suitable measure for the hardness of negative samples together with similarity by a designed probability estimator. Although verified the effectiveness, previous methods neglect the hard positive sample pair, thus leading to sub-optimal performance. In this work, we argue that the samples with the same category but low similarity should also be carefully learned. From this motivation, we propose a general sample weighting strategy to guide the network focus more on both the hard positive and negative samples.




\begin{figure*}
\centering
\includegraphics[scale=0.56]{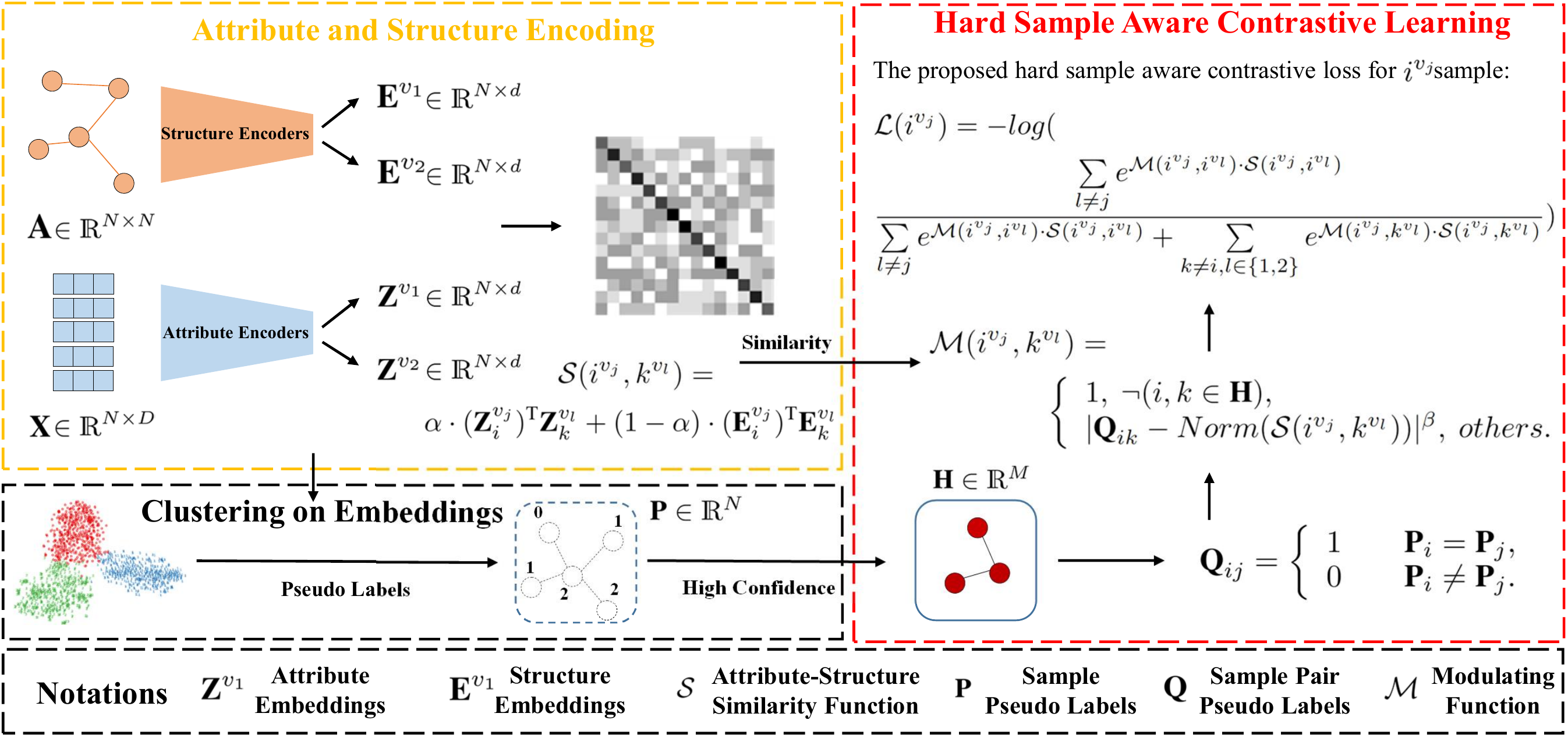}
\caption{Illustration of our proposed hard sample aware network. In attribute and structure encoding, we embed the attribute and structure into the latent space with the attribute encoders and structure encoders. Then the sample similarities are calculated by a learnable linear combination of attribute similarity and structure similarity, thus better revealing the sample relations. Moreover, guided by the high-confidence information, a general dynamic sample weighting strategy is proposed to up-weight hard sample pairs while down-weighting the easy ones. Overall, the hard sample aware contrastive loss guides the network to focus more on both hard positive and negative sample pairs, thus further improving the discriminative capability of samples.}
\label{OVERRALL_FIGURE}  
\end{figure*}




\section{Method}
In this section, we propose a novel Hard Sample Aware Network (HSAN) for contrastive deep graph clustering by guiding our network to focus on both the hard positive and negative sample pairs. The framework is shown in Figure \ref{OVERRALL_FIGURE}.


\subsection{Notation and Problem Definition}
Let $\mathcal{V}=\{v_1, v_2, \dots, v_N\}$ be a set of $N$ nodes with $C$ classes and $\mathcal{E}$ be a set of edges. In the matrix form, $\textbf{X} \in \mathds{R}^{N\times D}$ and $\textbf{A} \in \mathds{R}^{N\times N}$ denotes the attribute matrix and the original adjacency matrix, respectively. Then $\mathcal{G}=\left \{\textbf{X}, \textbf{A} \right \}$ denotes an undirected graph. The degree matrix is formulated as $\textbf{D}=diag(d_1, d_2, \dots ,d_N)\in \mathds{R}^{N\times N}$ and $d_i=\sum_{(v_i,v_j)\in \mathcal{E}}a_{ij}$. The graph Laplacian matrix is defined as $\textbf{L}=\textbf{D}-\textbf{A}$. With the renormalization trick $\widehat{\textbf{A}}=\textbf{A}+\textbf{I}$ in GCN \cite{GCN}, the symmetric normalized graph Laplacian matrix is denoted as $\widetilde{\textbf{L}} = \textbf{I}-\widehat{\textbf{D}}^{-\frac{1}{2}}\widehat{\textbf{A}}\widehat{\textbf{D}}^{-\frac{1}{2}}$. The notations are summarized in Table \ref{NOTATION_TABLE}.

The target of deep graph clustering is to encode nodes with the neural network in an unsupervised manner and then divide them into several disjoint groups. In general, a neural network $\mathcal{F}$ is firstly trained without human annotations and embeds the nodes into the latent space by exploiting the node attributes and the graph structure as follows:

\begin{equation} 
\textbf{E} = \mathcal{F}(\textbf{A}, \textbf{X}),
\label{Embed}
\end{equation}
where \textbf{X} and \textbf{A} denotes the attribute matrix and the original adjacency matrix. Besides, $\textbf{E} \in \mathds{R}^{N \times d}$ is the learned node embeddings, where $N$ is the number of samples and $d$ is the number of feature dimensions. After that, a clustering algorithm $\mathcal{C}$ such as K-means, 
spectral clustering, or the clustering neural network layer \cite{SDCN} is adopted to divide nodes into $K$ disjoint groups as follows:

\begin{equation} 
\Phi = \mathcal{C}(\textbf{E}),
\label{clustering}
\end{equation}
where $\Phi \in \mathds{R}^{N \times K}$ denotes the cluster membership matrix for all $N$ samples.

\begin{table}[!t]
\centering
\small
\scalebox{1.0}{
\begin{tabular}{ll}
\toprule
\textbf{Notation}                                        & \textbf{Meaning}                                \\ \midrule
$\textbf{X}\in \mathds{R}^{N\times D}$  & Attribute matrix  
\\
$\widetilde{\textbf{X}}\in \mathds{R}^{N\times D}$  & Low-pass filtered attribute matrix
\\
$\textbf{A}\in \mathds{R}^{N\times N}$  & Original adjacency matrix   
\\
$\widehat{\textbf{A}}\in \mathds{R}^{N\times N}$  & Adjacency matrix with self-loop
\\
$\textbf{L}\in \mathds{R}^{N\times N}$  & Graph Laplacian matrix
\\
$\widetilde{\textbf{L}}\in \mathds{R}^{N\times N}$  & Symmetric normalized Laplacian matrix
\\
$\textbf{Z}^{v_k} \in \mathds{R}^{N\times d}$   & Attribute embeddings in $k$-th view      
\\
$\textbf{E}^{v_k} \in \mathds{R}^{N\times d}$  & Structure embeddings in $k$-th view
\\
$\textbf{P} \in \mathds{R}^{N}$ & Sample clustering pseudo labels
\\
$\textbf{H}\in \mathds{R}^{M}$ & High-confidence sample set
\\
$\textbf{Q} \in \mathds{R}^{N \times N}$ & Sample pair pseudo labels
\\
$\| \cdot \|_2$  & L-2 norm
\\
\bottomrule
\end{tabular}
}
\caption{Notation summary.}
\label{NOTATION_TABLE} 
\end{table}

\subsection{Attribute and Structure Encoding}
In this section, we design two types of encoders to embed the nodes into the latent space. Concretely, the attribute encoder (AE) and the structure encoder (SE) encode the attribute and structural information of samples, respectively.

In the process of attribute encoding, we follow previous work \cite{AGE} and filter the high-frequency noises in the attribute matrix $\widetilde{\textbf{X}}$ as formulated:

\begin{equation}
\widetilde{\textbf{X}} = (\prod_{i=1}^{t}(\textbf{I} - \widetilde{\textbf{L}}))\textbf{X} = (\textbf{I} - \widetilde{\textbf{L}})^t\textbf{X},
\label{SMOOTH}
\end{equation}
where $\textbf{I} - \widetilde{\textbf{L}}$ is the graph Laplacian
filter and $t$ is the filtering times. Then we encode $\widetilde{\textbf{X}}$ with $\text{AE}_1$ and $\text{AE}_2$ as follows:

\begin{equation} 
\begin{aligned}
\textbf{Z}^{v_1} &= \text{AE}_1(\widetilde{\textbf{X}}); \textbf{Z}^{v_1}_i = \frac{\textbf{Z}^{v_1}_i}{||\textbf{Z}^{v_1}_i||_2}, i=1,2,...,N; \\
\textbf{Z}^{v_2} &= \text{AE}_2(\widetilde{\textbf{X}});  \textbf{Z}^{v_2}_j = \frac{\textbf{Z}^{v_2}_j}{||\textbf{Z}^{v_2}_j||_2}, j=1,2,...,N,
\end{aligned}
\label{AE}
\end{equation}
where $\textbf{Z}^{v_1}$ and $\textbf{Z}^{v_2}$ denote two-view attribute embeddings of the samples. Here, $\text{AE}_1$ and $\text{AE}_2$ are both simple multi-layer perceptions (MLPs), which have the same architecture but un-shared parameters, thus $\textbf{Z}^{v_1}$ and $\textbf{Z}^{v_2}$ contain different semantic information. Different the mixup-based methods \cite{xihong_MGCN,lirong_graph_mixup}, our proposed method is free from augmentation.

In addition, we further propose the structure encoder to encode the structural information of samples. Concretely, we design the structure encoder as follows:
\begin{equation} 
\begin{aligned}
\textbf{E}^{v_1} &= \text{SE}_1(\textbf{A}); \textbf{E}^{v_1}_i = \frac{\textbf{E}^{v_1}_i}{||\textbf{E}^{v_1}_i||_2}, i=1,2,...,N; \\
\textbf{E}^{v_2} &= \text{SE}_2(\textbf{A});  \textbf{E}^{v_2}_j = \frac{\textbf{E}^{v_2}_j}{||\textbf{E}^{v_2}_j||_2},j=1,2,...,N,
\end{aligned}
\label{SE}
\end{equation}
where $\textbf{E}^{v_1}$ and $\textbf{E}^{v_2}$ denote two-view structure embeddings. Similarly, $\text{SE}_1$ and $\text{SE}_2$ are simple MLPs, which have the same architecture but un-shared parameters, thus $\textbf{E}^{v_1}$ and $\textbf{E}^{v_2}$ contain different semantic information during training.

In this manner, we obtain the attribute embedding and structure embedding of each sample. Subsequently, the attribute-structure similarity function $\mathcal{S}$ is proposed to calculate the similarity between $i$-th sample in the $j$-th view and $k$-th sample in the $l$-th view as formulated:
\begin{equation} 
\begin{aligned}
\mathcal{S}(i^{v_j},k^{v_l}) =  \alpha \cdot (\textbf{Z}_i^{v_j})^\text{T} \textbf{Z}_k^{v_l}  +  (1-\alpha) \cdot (\textbf{E}_i^{v_j})^\text{T} \textbf{E}_k^{v_l},
\end{aligned}
\label{SIM}
\end{equation}
where $i,k \in \{1, 2, ..., N\}$ and $j,l \in \{1,2\}$. Besides, $\alpha$ denotes a learnable trade-off parameter. The first and second term in Eq. \eqref{SIM} both denotes the cosine similarity. $\mathcal{S}$ can better reveal sample relations by considering both attribute and structure information, thus assisting the hard sample mining.


\subsection{Clustering and Pseudo Label Generation}
After encoding, K-means is performed on the learned node embeddings to obtain the clustering results. Then we extract the more reliable clustering information as follows. To be specific, we first generate the clustering pseudo labels $\textbf{P} \in \mathds{R}^{N}$ and then select the top $\tau$ samples as the high-confidence sample set $\textbf{H}\in \mathds{R}^{M}$. Here, $\tau$ is the confidence hyper-parameter and $M$ is the number of high-confidence samples. The confidence is measured by the distance to the cluster center \cite{IDCRN}. Based on $\textbf{P}$, we calculate the sample pair pseudo labels $\textbf{Q} \in \mathds{R}^{N \times N}$ as follows:

\begin{equation} 
\textbf{Q}_{ij}=\left\{
\begin{array}{rcl}
1       &      & {\textbf{P}_i = \textbf{P}_j}, \\
0     &      & {\textbf{P}_i\neq \textbf{P}_j}.
\end{array} \right.
\label{Q_generation}
\end{equation}
Here, $\textbf{Q}_{ij}$ reveals the pseudo relation between $i$-th and $j$-th samples. Precisely, $\textbf{Q}_{ij}=1$ means $i-th$ and $j$-th samples are more likely to be the positive sample pair while $\textbf{Q}_{ij}=0$ implies they are more likely to be the negative ones.

\subsection{Hard Sample Aware Contrastive Learning}
In this section, we firstly introduce the drawback of classical infoNCE loss \cite{infoNCE} in the graph contrastive methods \cite{GRACE,ProGCL} and then propose a novel hard sample aware contrastive loss to guide our network to focus more on hard positive and negative samples.

The classical infoNCE loss for $i$-th sample in the $j$-th view is formulated as follows:
\begin{equation} 
\begin{aligned}
&\mathcal{L}_{infoNCE}(i^{v_j}) = \\& -log\frac{e^{\theta(i^{v_j}, i^{v_l})}}{e^{\theta(i^{v_j}, i^{v_l})}+\sum\limits_{k\neq i}({e^{\theta (i^{v_j},k^{v_j})}}+e^{\theta (i^{v_j}, k^{v_l})})}.
\end{aligned}
\label{infoNCE}
\end{equation}
where $j\neq l$. Besides, $\theta(\cdot)$ denotes the cosine similarity between the paired samples in the latent space. By minimizing infoNCE loss, they pull together the same samples in different views while pushing away other samples.

However, we find the drawback of classical infoNCE is that the hard sample pairs are treated equally to the easy ones, limiting the discriminative capability of the network. To solve this problem, we propose a weight modulating function $\mathcal{M}$ to dynamically adjust the weights of sample pairs during training. Concretely, based on the proposed attribute-structure similarity function $\mathcal{S}$ and the generated sample pair pseudo labels $\textbf{Q}$, $\mathcal{M}$ is formulated as follows:



\begin{equation} 
\mathcal{M}(i^{v_j},k^{v_l})=\left\{
\begin{array}{l} 1, \ \neg(i,k\in\textbf{H}), \\ 

|\textbf{Q}_{ik}-Norm(\mathcal{S}(i^{v_j},k^{v_l}))|^{\beta}, \  others.
\end{array} \right.
\label{M}
\end{equation}
where $i^{v_j}$ denotes the $i$-th samples in the $j$-th view and $Norm$ denotes the min-max normalization. In Eq. \eqref{M}, when $i$-th or $j$-th sample is not with high-confidence, i.e., $\neg(i,k\in\textbf{H})$, we keep the original setting in the infoNCE loss. Differently, the sample weights are modulated with the pseudo information and the similarity of samples, when the samples are with high-confidence. Here, the hyper-parameter $\beta \in [1,\ 5]$ is the focusing factor, which determines the down-weighting rate of the easy sample pairs. In the following, we analyze the properties of the proposed weight modulating function $\mathcal{M}$. 


1) $\mathcal{M}$ can up-weight the hard samples while down-weighting the easy samples. Concretely, when $i$-th, $j$-th samples are recognized as positive sample pair ($\textbf{Q}_{ij}=1$), the hardness of pulling them together decreases as the similarity increases. Thus, $\mathcal{M}$ up-weights the positive sample pairs with small similarity (the hard ones) while down-weighting that with large similarity (the easy ones). For example, when $\textbf{Q}_{ij}=1$ and $\beta=2$, the easy positive sample pair with $0.90$ similarity is weighted with $0.01$. Differently, the hard positive sample pair with $0.10$ similarity is weighted with $0.81$, which is significantly larger than $0.01$. The similar conclusion can be deduced on the negative sample pairs. Experimental evidence can be found in Figure \ref{motivation}. 

2) The focusing factor $\beta$ controls the down-weighting rate of easy sample pairs. Concretely, when $\beta$ increases, the down-weighting rate of easy sample pairs increases and vice versa. Take the positive sample pair as an example ($\textbf{Q}_{ij}=1$), the easy ones with $0.9$ similarity be down-weighted to $0.1^{\beta}$. Here, when $\beta=1$, the weight is set to $0.1$. Differently, when $\beta=3$, the weight is degraded to $0.001$, which is obviously smaller than $0.1$. This property is verified by visualization experiments in Figure 2-3 of Appendix.

Based on $\mathcal{S}$ and $\mathcal{M}$, we formulate the hard sample aware contrastive loss for $i$-th sample in $j$-th view as follows:

\begin{equation} 
\begin{aligned}
&\mathcal{L}(i^{v_j})=-log( \\ &\frac{\sum\limits_{l \neq j} e^{\mathcal{M}(i^{v_j},i^{v_l}) \cdot \mathcal{S}(i^{v_j},i^{v_l})}}{\sum\limits_{l \neq j} e^{\mathcal{M}(i^{v_j},i^{v_l}) \cdot \mathcal{S}(i^{v_j},i^{v_l})}+\sum\limits_{k\neq i,l \in \{1,2\}}{e^{\mathcal{M}(i^{v_j},k^{v_l}) \cdot \mathcal{S}(i^{v_j},k^{v_l})}}}).
\end{aligned}
\label{our_loss_part}
\end{equation}
Compared with the classical infoNCE loss, we first adopt a more comprehensive similarity measure criterion $\mathcal{S}$ to assist sample hardness measurement. Then we propose a weight modulating function $\mathcal{M}$ to up-weight hard sample pairs and down-weight the easy ones. In summary, the overall loss of our method is formulated as follows:
\begin{equation} 
\mathcal{L}=\frac{1}{2N}\sum_{j=1}^2\sum_{i=1}^N \mathcal{L}(i^{v_j}).
\label{our_loss}
\end{equation} 

This hard sample aware contrastive loss can guide our network to focus on not only the hard negative samples but also the hard positive ones, thus improving the discriminative capability of samples further. We summarize two reasons as follows. 1) The proposed attribute-structure similarity function $\mathcal{S}$ considers attribute and structure information, thus better revealing the sample relations. 2) The proposed weight modulating function $\mathcal{M}$ is a general dynamic sample weighting strategy for positive and negative sample pairs. It can up-weight the hard sample pairs while down-weighting the easy ones during training.







\subsection{Complexity Analysis of Loss Function}
In this section, we analyze the time and space complexity of the proposed hard sample aware contrastive loss $\mathcal{L}$. Here, we denote the batch size is $B$, the number of the high confident samples in this batch is $M$, and the dimension of embeddings is $d$. The time complexity of $\mathcal{S}$ and $\mathcal{M}$ is $\mathcal{O}(B^2d)$ and $\mathcal{O}(M^2d)$, respectively. Since $M<B$, the time complexity of the whole hard sample aware contrastive loss is $\mathcal{O}(B^2d)$. Besides, the space complexity of our proposed loss is $\mathcal{O}(B^2)$. Thus, the proposed loss will not bring the high time or space costs compared with the classical infoNCE loss. The detailed process of our proposed HSAN is summarized in Algorithm \ref{ALGORITHM}. Besides, the PyTorch-style pseudo code can be found in Appendix.

\begin{algorithm}[!t]
\small
\caption{Hard Sample Aware Network}
\label{ALGORITHM}
\flushleft{\textbf{Input}: Input graph $\mathcal{G}=\{\textbf{X},\textbf{A}\}$; cluster number $C$; epoch number $I$; filtering times $t$; confidence $\tau$; focusing factor $\beta$.} \\
\flushleft{$\textbf{Output}$: The clustering result $\Phi$.} 
\begin{algorithmic}[1]
\STATE Obtain the low-pass filtered attribute matrix $\widetilde{\textbf{X}}$ in Eq. \eqref{SMOOTH}.
\FOR{$i=1$ to $I$}
\STATE Encode $\widetilde{\textbf{X}}$ with attribute encoders $\text{AE}_1$ and $\text{AE}_2$ in Eq. \eqref{AE}.
\STATE Encode $\textbf{A}$ with structure encoders $\text{SE}_1$ and $\text{SE}_2$ in Eq. \eqref{SE}.
\STATE Calculate the attribute-structure similarity $\mathcal{S}$ by Eq. \eqref{SIM}.
\STATE Perform K-means on node embeddings and obtain high-confidence sample pair pseudo labels $\textbf{Q}$ in Eq. \eqref{Q_generation}.
\STATE Calculate the modulated sample weights in Eq. \eqref{M}.
\STATE Update model by minimizing the hard sample aware contrastive loss $\mathcal{L}$ in Eq. \eqref{our_loss}.
\ENDFOR
\STATE{Obtain $\Phi$ by performing K-means over the node embeddings.}
\STATE \textbf{return} $\Phi$
\end{algorithmic}
\end{algorithm}

\section{Experiments}
\subsection{Dataset}
To evaluate the effectiveness of our proposed HSAN, we conduct experiments on six benchmark datasets, including CORA, CITE, Amazon Photo (AMAP), Brazil Air-Traffic (BAT), Europe Air-Traffic (EAT), and USA Air-Traffic (UAT). The brief information of datasets is summarized in Table 1 of the Appendix.


\begin{table*}[!t]
\centering
\setlength{\tabcolsep}{4pt}
\scalebox{0.60}{
\begin{tabular}{c|c|ccccc|cccccc|ccc}
\hline
{\color[HTML]{000000} }                                   & {\color[HTML]{000000} }                                  & \multicolumn{5}{c|}{{\color[HTML]{000000} \textbf{Classical Deep Graph Clustering}}}                                                                                                                                                & \multicolumn{6}{c|}{{\color[HTML]{000000} \textbf{Contrastive Deep Graph Clustering}}}                                                                                                                                                                                                           & \multicolumn{3}{c}{{\color[HTML]{000000} \textbf{Hard Sample Mining}}}                                                                  \\ \cline{3-16} 
{\color[HTML]{000000} }                                   & {\color[HTML]{000000} }                                  & {\color[HTML]{000000} \textbf{MGAE}}       & {\color[HTML]{000000} \textbf{DAEGC}}        & {\color[HTML]{000000} \textbf{ARGA}}         & {\color[HTML]{000000} \textbf{SDCN}}       & {\color[HTML]{000000} \textbf{DFCN}}        & {\color[HTML]{000000} \textbf{AGE}}           & {\color[HTML]{000000} \textbf{MVGRL}}       & {\color[HTML]{000000} \textbf{AutoSSL}}     & {\color[HTML]{000000} \textbf{AGC-DRR}}      & {\color[HTML]{000000} \textbf{DCRN}}                  & {\color[HTML]{000000} \textbf{AFGRL}}       & {\color[HTML]{000000} \textbf{GDCL}}         & {\color[HTML]{000000} \textbf{ProGCL}}      & {\color[HTML]{000000} \textbf{HSAN}}        \\
\multirow{-3}{*}{{\color[HTML]{000000} \textbf{Dataset}}} & \multirow{-3}{*}{{\color[HTML]{000000} \textbf{Metric}}} & {\color[HTML]{000000} \textbf{CIKM 19}}  & {\color[HTML]{000000} \textbf{IJCAI   19}} & {\color[HTML]{000000} \textbf{IJCAI   19}} & {\color[HTML]{000000} \textbf{WWW   20}} & {\color[HTML]{000000} \textbf{AAAI   21}} & {\color[HTML]{000000} \textbf{SIGKDD   20}} & {\color[HTML]{000000} \textbf{ICML   20}} & {\color[HTML]{000000} \textbf{ICLR   22}} & {\color[HTML]{000000} \textbf{IJCAI   22}} & {\color[HTML]{000000} \textbf{AAAI   22}}           & {\color[HTML]{000000} \textbf{AAAI   22}} & {\color[HTML]{000000} \textbf{IJCAI   21}} & {\color[HTML]{000000} \textbf{ICML   22}} & {\color[HTML]{000000} \textbf{Ours}}       \\ \hline
{\color[HTML]{000000} }                                   & {\color[HTML]{000000} \textbf{ACC}}                      & {\color[HTML]{000000} \textbf{43.38±2.11}} & {\color[HTML]{000000} \textbf{70.43±0.36}}   & {\color[HTML]{000000} \textbf{71.04±0.25}}   & {\color[HTML]{000000} \textbf{35.60±2.83}} & {\color[HTML]{000000} \textbf{36.33±0.49}}  & {\color[HTML]{0000FF} \textbf{73.50±1.83}}    & {\color[HTML]{000000} \textbf{70.47±3.70}}  & {\color[HTML]{000000} \textbf{63.81±0.57}}  & {\color[HTML]{000000} \textbf{40.62±0.55}}   & {\color[HTML]{000000} \textbf{61.93±0.47}}            & {\color[HTML]{000000} \textbf{26.25±1.24}}  & {\color[HTML]{000000} \textbf{70.83±0.47}}   & {\color[HTML]{000000} \textbf{57.13±1.23}}  & {\color[HTML]{FF0000} \textbf{77.07±1.56}} \\
{\color[HTML]{000000} }                                   & {\color[HTML]{000000} \textbf{NMI}}                      & {\color[HTML]{000000} \textbf{28.78±2.97}} & {\color[HTML]{000000} \textbf{52.89±0.69}}   & {\color[HTML]{000000} \textbf{51.06±0.52}}   & {\color[HTML]{000000} \textbf{14.28±1.91}} & {\color[HTML]{000000} \textbf{19.36±0.87}}  & {\color[HTML]{0000FF} \textbf{57.58±1.42}}    & {\color[HTML]{000000} \textbf{55.57±1.54}}  & {\color[HTML]{000000} \textbf{47.62±0.45}}  & {\color[HTML]{000000} \textbf{18.74±0.73}}   & {\color[HTML]{000000} \textbf{45.13±1.57}}            & {\color[HTML]{000000} \textbf{12.36±1.54}}  & {\color[HTML]{000000} \textbf{56.60±0.36}}   & {\color[HTML]{000000} \textbf{41.02±1.34}}  & {\color[HTML]{FF0000} \textbf{59.21±1.03}} \\
{\color[HTML]{000000} }                                   & {\color[HTML]{000000} \textbf{ARI}}                      & {\color[HTML]{000000} \textbf{16.43±1.65}} & {\color[HTML]{000000} \textbf{49.63±0.43}}   & {\color[HTML]{000000} \textbf{47.71±0.33}}   & {\color[HTML]{000000} \textbf{07.78±3.24}} & {\color[HTML]{000000} \textbf{04.67±2.10}}  & {\color[HTML]{0000FF} \textbf{50.10±2.14}}    & {\color[HTML]{000000} \textbf{48.70±3.94}}  & {\color[HTML]{000000} \textbf{38.92±0.77}}  & {\color[HTML]{000000} \textbf{14.80±1.64}}   & {\color[HTML]{000000} \textbf{33.15±0.14}}            & {\color[HTML]{000000} \textbf{14.32±1.87}}  & {\color[HTML]{000000} \textbf{48.05±0.72}}   & {\color[HTML]{000000} \textbf{30.71±2.70}}  & {\color[HTML]{FF0000} \textbf{57.52±2.70}} \\
\multirow{-4}{*}{{\color[HTML]{000000} \textbf{CORA}}}    & {\color[HTML]{000000} \textbf{F1}}                       & {\color[HTML]{000000} \textbf{33.48±3.05}} & {\color[HTML]{000000} \textbf{68.27±0.57}}   & {\color[HTML]{000000} \textbf{69.27±0.39}}   & {\color[HTML]{000000} \textbf{24.37±1.04}} & {\color[HTML]{000000} \textbf{26.16±0.50}}  & {\color[HTML]{0000FF} \textbf{69.28±1.59}}    & {\color[HTML]{000000} \textbf{67.15±1.86}}  & {\color[HTML]{000000} \textbf{56.42±0.21}}  & {\color[HTML]{000000} \textbf{31.23±0.57}}   & {\color[HTML]{000000} \textbf{49.50±0.42}}            & {\color[HTML]{000000} \textbf{30.20±1.15}}  & {\color[HTML]{000000} \textbf{52.88±0.97}}   & {\color[HTML]{000000} \textbf{45.68±1.29}}  & {\color[HTML]{FF0000} \textbf{75.11±1.40}} \\ \hline
{\color[HTML]{000000} }                                   & {\color[HTML]{000000} \textbf{ACC}}                      & {\color[HTML]{000000} \textbf{61.35±0.80}} & {\color[HTML]{000000} \textbf{64.54±1.39}}   & {\color[HTML]{000000} \textbf{61.07±0.49}}   & {\color[HTML]{000000} \textbf{65.96±0.31}} & {\color[HTML]{000000} \textbf{69.50±0.20}}  & {\color[HTML]{000000} \textbf{69.73±0.24}}    & {\color[HTML]{000000} \textbf{62.83±1.59}}  & {\color[HTML]{000000} \textbf{66.76±0.67}}  & {\color[HTML]{000000} \textbf{68.32±1.83}}   & {\color[HTML]{0000FF} \textbf{70.86±0.18}}            & {\color[HTML]{000000} \textbf{31.45±0.54}}  & {\color[HTML]{000000} \textbf{66.39±0.65}}   & {\color[HTML]{000000} \textbf{65.92±0.80}}  & {\color[HTML]{FF0000} \textbf{71.15±0.80}} \\
{\color[HTML]{000000} }                                   & {\color[HTML]{000000} \textbf{NMI}}                      & {\color[HTML]{000000} \textbf{34.63±0.65}} & {\color[HTML]{000000} \textbf{36.41±0.86}}   & {\color[HTML]{000000} \textbf{34.40±0.71}}   & {\color[HTML]{000000} \textbf{38.71±0.32}} & {\color[HTML]{000000} \textbf{43.90±0.20}}  & {\color[HTML]{000000} \textbf{44.93±0.53}}    & {\color[HTML]{000000} \textbf{40.69±0.93}}  & {\color[HTML]{000000} \textbf{40.67±0.84}}  & {\color[HTML]{000000} \textbf{43.28±1.41}}   & {\color[HTML]{FF0000} \textbf{45.86±0.35}}            & {\color[HTML]{000000} \textbf{15.17±0.47}}  & {\color[HTML]{000000} \textbf{39.52±0.38}}   & {\color[HTML]{000000} \textbf{39.59±0.39}}  & {\color[HTML]{0000FF} \textbf{45.06±0.74}} \\
{\color[HTML]{000000} }                                   & {\color[HTML]{000000} \textbf{ARI}}                      & {\color[HTML]{000000} \textbf{33.55±1.18}} & {\color[HTML]{000000} \textbf{37.78±1.24}}   & {\color[HTML]{000000} \textbf{34.32±0.70}}   & {\color[HTML]{000000} \textbf{40.17±0.43}} & {\color[HTML]{000000} \textbf{45.50±0.30}}  & {\color[HTML]{000000} \textbf{45.31±0.41}}    & {\color[HTML]{000000} \textbf{34.18±1.73}}  & {\color[HTML]{000000} \textbf{38.73±0.55}}  & {\color[HTML]{000000} \textbf{45.34±2.33}}   & {\color[HTML]{FF0000} \textbf{47.64±0.30}}            & {\color[HTML]{000000} \textbf{14.32±0.78}}  & {\color[HTML]{000000} \textbf{41.07±0.96}}   & {\color[HTML]{000000} \textbf{36.16±1.11}}  & {\color[HTML]{0000FF} \textbf{47.05±1.12}} \\
\multirow{-4}{*}{{\color[HTML]{000000} \textbf{CITE}}}    & {\color[HTML]{000000} \textbf{F1}}                       & {\color[HTML]{000000} \textbf{57.36±0.82}} & {\color[HTML]{000000} \textbf{62.20±1.32}}   & {\color[HTML]{000000} \textbf{58.23±0.31}}   & {\color[HTML]{000000} \textbf{63.62±0.24}} & {\color[HTML]{000000} \textbf{64.30±0.20}}  & {\color[HTML]{000000} \textbf{64.45±0.27}}    & {\color[HTML]{000000} \textbf{59.54±2.17}}  & {\color[HTML]{000000} \textbf{58.22±0.68}}  & {\color[HTML]{0000FF} \textbf{64.82±1.60}}   & {\color[HTML]{FF0000} \textbf{65.83±0.21}}            & {\color[HTML]{000000} \textbf{30.20±0.71}}  & {\color[HTML]{000000} \textbf{61.12±0.70}}   & {\color[HTML]{000000} \textbf{57.89±1.98}}  & {\color[HTML]{000000} \textbf{63.01±1.79}} \\ \hline
{\color[HTML]{000000} }                                   & {\color[HTML]{000000} \textbf{ACC}}                      & {\color[HTML]{000000} \textbf{71.57±2.48}} & {\color[HTML]{000000} \textbf{75.96±0.23}}   & {\color[HTML]{000000} \textbf{69.28±2.30}}   & {\color[HTML]{000000} \textbf{53.44±0.81}} & {\color[HTML]{0000FF} \textbf{76.82±0.23}}  & {\color[HTML]{000000} \textbf{75.98±0.68}}    & {\color[HTML]{000000} \textbf{41.07±3.12}}  & {\color[HTML]{000000} \textbf{54.55±0.97}}  & {\color[HTML]{000000} \textbf{76.81±1.45}}   & {\color[HTML]{000000} }                               & {\color[HTML]{000000} \textbf{75.51±0.77}}  & {\color[HTML]{000000} \textbf{43.75±0.78}}   & {\color[HTML]{000000} \textbf{51.53±0.38}}  & {\color[HTML]{FF0000} \textbf{77.02±0.33}} \\
{\color[HTML]{000000} }                                   & {\color[HTML]{000000} \textbf{NMI}}                      & {\color[HTML]{000000} \textbf{62.13±2.79}} & {\color[HTML]{000000} \textbf{65.25±0.45}}   & {\color[HTML]{000000} \textbf{58.36±2.76}}   & {\color[HTML]{000000} \textbf{44.85±0.83}} & {\color[HTML]{000000} \textbf{66.23±1.21}}  & {\color[HTML]{000000} \textbf{65.38±0.61}}    & {\color[HTML]{000000} \textbf{30.28±3.94}}  & {\color[HTML]{000000} \textbf{48.56±0.71}}  & {\color[HTML]{0000FF} \textbf{66.54±1.24}}   & {\color[HTML]{000000} }                               & {\color[HTML]{000000} \textbf{64.05±0.15}}  & {\color[HTML]{000000} \textbf{37.32±0.28}}   & {\color[HTML]{000000} \textbf{39.56±0.39}}  & {\color[HTML]{FF0000} \textbf{67.21±0.33}} \\
{\color[HTML]{000000} }                                   & {\color[HTML]{000000} \textbf{ARI}}                      & {\color[HTML]{000000} \textbf{48.82±4.57}} & {\color[HTML]{000000} \textbf{58.12±0.24}}   & {\color[HTML]{000000} \textbf{44.18±4.41}}   & {\color[HTML]{000000} \textbf{31.21±1.23}} & {\color[HTML]{000000} \textbf{58.28±0.74}}  & {\color[HTML]{000000} \textbf{55.89±1.34}}    & {\color[HTML]{000000} \textbf{18.77±2.34}}  & {\color[HTML]{000000} \textbf{26.87±0.34}}  & {\color[HTML]{FF0000} \textbf{60.15±1.56}}   & {\color[HTML]{000000} }                               & {\color[HTML]{000000} \textbf{54.45±0.48}}  & {\color[HTML]{000000} \textbf{21.57±0.51}}   & {\color[HTML]{000000} \textbf{34.18±0.89}}  & {\color[HTML]{0000FF} \textbf{58.01±0.48}} \\
\multirow{-4}{*}{{\color[HTML]{000000} \textbf{AMAP}}}    & {\color[HTML]{000000} \textbf{F1}}                       & {\color[HTML]{000000} \textbf{68.08±1.76}} & {\color[HTML]{000000} \textbf{69.87±0.54}}   & {\color[HTML]{000000} \textbf{64.30±1.95}}   & {\color[HTML]{000000} \textbf{50.66±1.49}} & {\color[HTML]{000000} \textbf{71.25±0.31}}  & {\color[HTML]{0000FF} \textbf{71.74±0.93}}    & {\color[HTML]{000000} \textbf{32.88±5.50}}  & {\color[HTML]{000000} \textbf{54.47±0.83}}  & {\color[HTML]{000000} \textbf{71.03±0.64}}   & \multirow{-4}{*}{{\color[HTML]{000000} \textbf{OOM}}} & {\color[HTML]{000000} \textbf{69.99±0.34}}  & {\color[HTML]{000000} \textbf{38.37±0.29}}   & {\color[HTML]{000000} \textbf{31.97±0.44}}  & {\color[HTML]{FF0000} \textbf{72.03±0.46}} \\ \hline
{\color[HTML]{000000} }                                   & {\color[HTML]{000000} \textbf{ACC}}                      & {\color[HTML]{000000} \textbf{53.59±2.04}} & {\color[HTML]{000000} \textbf{52.67±0.00}}   & {\color[HTML]{0000FF} \textbf{67.86±0.80}}   & {\color[HTML]{000000} \textbf{53.05±4.63}} & {\color[HTML]{000000} \textbf{55.73±0.06}}  & {\color[HTML]{000000} \textbf{56.68±0.76}}    & {\color[HTML]{000000} \textbf{37.56±0.32}}  & {\color[HTML]{000000} \textbf{42.43±0.47}}  & {\color[HTML]{000000} \textbf{47.79±0.02}}   & {\color[HTML]{000000} \textbf{67.94±1.45}}            & {\color[HTML]{000000} \textbf{50.92±0.44}}  & {\color[HTML]{000000} \textbf{45.42±0.54}}   & {\color[HTML]{000000} \textbf{55.73±0.79}}  & {\color[HTML]{FF0000} \textbf{77.15±0.72}} \\
{\color[HTML]{000000} }                                   & {\color[HTML]{000000} \textbf{NMI}}                      & {\color[HTML]{000000} \textbf{30.59±2.06}} & {\color[HTML]{000000} \textbf{21.43±0.35}}   & {\color[HTML]{0000FF} \textbf{49.09±0.54}}   & {\color[HTML]{000000} \textbf{25.74±5.71}} & {\color[HTML]{000000} \textbf{48.77±0.51}}  & {\color[HTML]{000000} \textbf{36.04±1.54}}    & {\color[HTML]{000000} \textbf{29.33±0.70}}  & {\color[HTML]{000000} \textbf{17.84±0.98}}  & {\color[HTML]{000000} \textbf{19.91±0.24}}   & {\color[HTML]{000000} \textbf{47.23±0.74}}            & {\color[HTML]{000000} \textbf{27.55±0.62}}  & {\color[HTML]{000000} \textbf{31.70±0.42}}   & {\color[HTML]{000000} \textbf{28.69±0.92}}  & {\color[HTML]{FF0000} \textbf{53.21±0.93}} \\
{\color[HTML]{000000} }                                   & {\color[HTML]{000000} \textbf{ARI}}                      & {\color[HTML]{000000} \textbf{24.15±1.70}} & {\color[HTML]{000000} \textbf{18.18±0.29}}   & {\color[HTML]{0000FF} \textbf{42.02±1.21}}   & {\color[HTML]{000000} \textbf{21.04±4.97}} & {\color[HTML]{000000} \textbf{37.76±0.23}}  & {\color[HTML]{000000} \textbf{26.59±1.83}}    & {\color[HTML]{000000} \textbf{13.45±0.03}}  & {\color[HTML]{000000} \textbf{13.11±0.81}}  & {\color[HTML]{000000} \textbf{14.59±0.13}}   & {\color[HTML]{000000} \textbf{39.76±0.87}}            & {\color[HTML]{000000} \textbf{21.89±0.74}}  & {\color[HTML]{000000} \textbf{19.33±0.57}}   & {\color[HTML]{000000} \textbf{21.84±1.34}}  & {\color[HTML]{FF0000} \textbf{52.20±1.11}} \\
\multirow{-4}{*}{{\color[HTML]{000000} \textbf{BAT}}}     & {\color[HTML]{000000} \textbf{F1}}                       & {\color[HTML]{000000} \textbf{50.83±3.23}} & {\color[HTML]{000000} \textbf{52.23±0.03}}   & {\color[HTML]{000000} \textbf{67.02±1.15}}   & {\color[HTML]{000000} \textbf{46.45±5.90}} & {\color[HTML]{000000} \textbf{50.90±0.12}}  & {\color[HTML]{000000} \textbf{55.07±0.80}}    & {\color[HTML]{000000} \textbf{29.64±0.49}}  & {\color[HTML]{000000} \textbf{34.84±0.15}}  & {\color[HTML]{000000} \textbf{42.33±0.51}}   & {\color[HTML]{0000FF} \textbf{67.40±0.35}}            & {\color[HTML]{000000} \textbf{46.53±0.57}}  & {\color[HTML]{000000} \textbf{39.94±0.57}}   & {\color[HTML]{000000} \textbf{56.08±0.89}}  & {\color[HTML]{FF0000} \textbf{77.13±0.76}} \\ \hline
{\color[HTML]{000000} }                                   & {\color[HTML]{000000} \textbf{ACC}}                      & {\color[HTML]{000000} \textbf{44.61±2.10}} & {\color[HTML]{000000} \textbf{36.89±0.15}}   & {\color[HTML]{0000FF} \textbf{52.13±0.00}}   & {\color[HTML]{000000} \textbf{39.07±1.51}} & {\color[HTML]{000000} \textbf{49.37±0.19}}  & {\color[HTML]{000000} \textbf{47.26±0.32}}    & {\color[HTML]{000000} \textbf{32.88±0.71}}  & {\color[HTML]{000000} \textbf{31.33±0.52}}  & {\color[HTML]{000000} \textbf{37.37±0.11}}   & {\color[HTML]{000000} \textbf{50.88±0.55}}            & {\color[HTML]{000000} \textbf{37.42±1.24}}  & {\color[HTML]{000000} \textbf{33.46±0.18}}   & {\color[HTML]{000000} \textbf{43.36±0.87}}  & {\color[HTML]{FF0000} \textbf{56.69±0.34}} \\
{\color[HTML]{000000} }                                   & {\color[HTML]{000000} \textbf{NMI}}                      & {\color[HTML]{000000} \textbf{15.60±2.30}} & {\color[HTML]{000000} \textbf{05.57±0.06}}   & {\color[HTML]{000000} \textbf{22.48±1.21}}   & {\color[HTML]{000000} \textbf{08.83±2.54}} & {\color[HTML]{0000FF} \textbf{32.90±0.41}}  & {\color[HTML]{000000} \textbf{23.74±0.90}}    & {\color[HTML]{000000} \textbf{11.72±1.08}}  & {\color[HTML]{000000} \textbf{07.63±0.85}}  & {\color[HTML]{000000} \textbf{07.00±0.85}}   & {\color[HTML]{000000} \textbf{22.01±1.23}}            & {\color[HTML]{000000} \textbf{11.44±1.41}}  & {\color[HTML]{000000} \textbf{13.22±0.33}}   & {\color[HTML]{000000} \textbf{23.93±0.45}}  & {\color[HTML]{FF0000} \textbf{33.25±0.44}} \\
{\color[HTML]{000000} }                                   & {\color[HTML]{000000} \textbf{ARI}}                      & {\color[HTML]{000000} \textbf{13.40±1.26}} & {\color[HTML]{000000} \textbf{05.03±0.08}}   & {\color[HTML]{000000} \textbf{17.29±0.50}}   & {\color[HTML]{000000} \textbf{06.31±1.95}} & {\color[HTML]{0000FF} \textbf{23.25±0.18}}  & {\color[HTML]{000000} \textbf{16.57±0.46}}    & {\color[HTML]{000000} \textbf{04.68±1.30}}  & {\color[HTML]{000000} \textbf{02.13±0.67}}  & {\color[HTML]{000000} \textbf{04.88±0.91}}   & {\color[HTML]{000000} \textbf{18.13±0.85}}            & {\color[HTML]{000000} \textbf{06.57±1.73}}  & {\color[HTML]{000000} \textbf{04.31±0.29}}   & {\color[HTML]{000000} \textbf{15.03±0.98}}  & {\color[HTML]{FF0000} \textbf{26.85±0.59}} \\
\multirow{-4}{*}{{\color[HTML]{000000} \textbf{EAT}}}     & {\color[HTML]{000000} \textbf{F1}}                       & {\color[HTML]{000000} \textbf{43.08±3.26}} & {\color[HTML]{000000} \textbf{34.72±0.16}}   & {\color[HTML]{0000FF} \textbf{52.75±0.07}}   & {\color[HTML]{000000} \textbf{33.42±3.10}} & {\color[HTML]{000000} \textbf{42.95±0.04}}  & {\color[HTML]{000000} \textbf{45.54±0.40}}    & {\color[HTML]{000000} \textbf{25.35±0.75}}  & {\color[HTML]{000000} \textbf{21.82±0.98}}  & {\color[HTML]{000000} \textbf{35.20±0.17}}   & {\color[HTML]{000000} \textbf{47.06±0.66}}            & {\color[HTML]{000000} \textbf{30.53±1.47}}  & {\color[HTML]{000000} \textbf{25.02±0.21}}   & {\color[HTML]{000000} \textbf{42.54±0.45}}  & {\color[HTML]{FF0000} \textbf{57.26±0.28}} \\ \hline
{\color[HTML]{000000} }                                   & {\color[HTML]{000000} \textbf{ACC}}                      & {\color[HTML]{000000} \textbf{48.97±1.52}} & {\color[HTML]{000000} \textbf{52.29±0.49}}   & {\color[HTML]{000000} \textbf{49.31±0.15}}   & {\color[HTML]{000000} \textbf{52.25±1.91}} & {\color[HTML]{000000} \textbf{33.61±0.09}}  & {\color[HTML]{0000FF} \textbf{52.37±0.42}}    & {\color[HTML]{000000} \textbf{44.16±1.38}}  & {\color[HTML]{000000} \textbf{42.52±0.64}}  & {\color[HTML]{000000} \textbf{42.64±0.31}}   & {\color[HTML]{000000} \textbf{49.92±1.25}}            & {\color[HTML]{000000} \textbf{41.50±0.25}} & {\color[HTML]{000000} \textbf{48.70±0.06}}   & {\color[HTML]{000000} \textbf{45.38±0.58}}  & {\color[HTML]{FF0000} \textbf{56.04±0.67}} \\
{\color[HTML]{000000} }                                   & {\color[HTML]{000000} \textbf{NMI}}                      & {\color[HTML]{000000} \textbf{20.69±0.98}} & {\color[HTML]{000000} \textbf{21.33±0.44}}   & {\color[HTML]{000000} \textbf{25.44±0.31}}   & {\color[HTML]{000000} \textbf{21.61±1.26}} & {\color[HTML]{0000FF} \textbf{26.49±0.41}}  & {\color[HTML]{000000} \textbf{23.64±0.66}}    & {\color[HTML]{000000} \textbf{21.53±0.94}}  & {\color[HTML]{000000} \textbf{17.86±0.22}}  & {\color[HTML]{000000} \textbf{11.15±0.24}}   & {\color[HTML]{000000} \textbf{24.09±0.53}}            & {\color[HTML]{000000} \textbf{17.33±0.54}}  & {\color[HTML]{000000} \textbf{25.10±0.01}}   & {\color[HTML]{000000} \textbf{22.04±2.23}}  & {\color[HTML]{FF0000} \textbf{26.99±2.11}} \\
{\color[HTML]{000000} }                                   & {\color[HTML]{000000} \textbf{ARI}}                      & {\color[HTML]{000000} \textbf{18.33±1.79}} & {\color[HTML]{000000} \textbf{20.50±0.51}}   & {\color[HTML]{000000} \textbf{16.57±0.31}}   & {\color[HTML]{000000} \textbf{21.63±1.49}} & {\color[HTML]{000000} \textbf{11.87±0.23}}  & {\color[HTML]{000000} \textbf{20.39±0.70}}    & {\color[HTML]{000000} \textbf{17.12±1.46}}  & {\color[HTML]{000000} \textbf{13.13±0.71}}  & {\color[HTML]{000000} \textbf{09.50±0.25}}   & {\color[HTML]{000000} \textbf{17.17±0.69}}            & {\color[HTML]{000000} \textbf{13.62±0.57}}  & {\color[HTML]{0000FF} \textbf{21.76±0.01}}   & {\color[HTML]{000000} \textbf{14.74±1.99}}  & {\color[HTML]{FF0000} \textbf{25.22±1.96}} \\
\multirow{-4}{*}{{\color[HTML]{000000} \textbf{UAT}}}     & {\color[HTML]{000000} \textbf{F1}}                       & {\color[HTML]{000000} \textbf{47.95±1.52}} & {\color[HTML]{0000FF} \textbf{50.33±0.64}}   & {\color[HTML]{000000} \textbf{50.26±0.16}}   & {\color[HTML]{000000} \textbf{45.59±3.54}} & {\color[HTML]{000000} \textbf{25.79±0.29}}  & {\color[HTML]{000000} \textbf{50.15±0.73}}    & {\color[HTML]{000000} \textbf{39.44±2.19}}  & {\color[HTML]{000000} \textbf{34.94±0.87}}  & {\color[HTML]{000000} \textbf{35.18±0.32}}   & {\color[HTML]{000000} \textbf{44.81±0.87}}            & {\color[HTML]{000000} \textbf{36.52±0.89}}  & {\color[HTML]{000000} \textbf{45.69±0.08}}   & {\color[HTML]{000000} \textbf{39.30±1.82}}  & {\color[HTML]{FF0000} \textbf{54.20±1.84}} \\ \hline
\end{tabular}}
\caption{The average clustering performance of ten runs on six benchmark datasets. The performance is evaluated by four metrics with mean value and standard deviation. The {\color[HTML]{FF0000}red} and {\color[HTML]{0000FF}blue} values indicate the best and the runner-up results, respectively.}
\label{COMPARE_TABLE}
\end{table*}

\begin{figure*}[!t]
\footnotesize
\begin{minipage}{0.139\linewidth}
\centerline{\includegraphics[width=\textwidth]{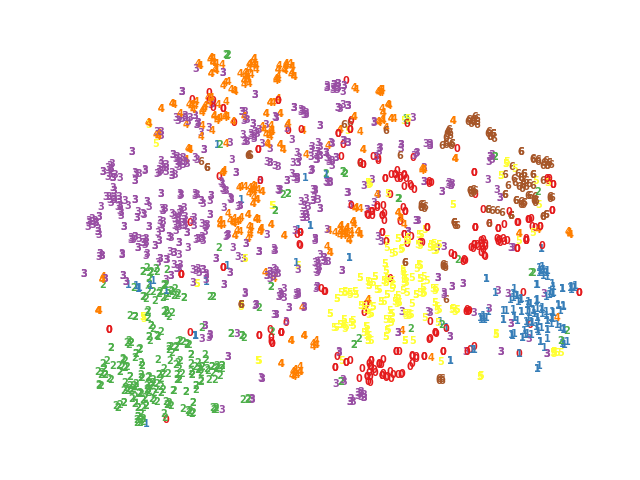}}
\vspace{3pt}
\centerline{\includegraphics[width=\textwidth]{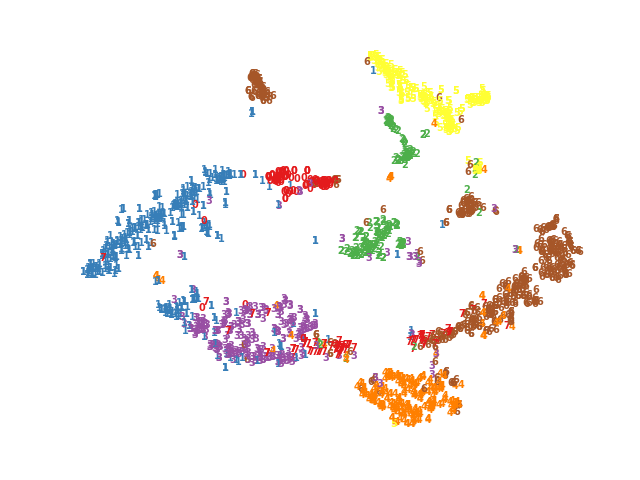}}
\vspace{3pt}
\centerline{(a) DAEGC}
\end{minipage}
\begin{minipage}{0.139\linewidth}
\centerline{\includegraphics[width=\textwidth]{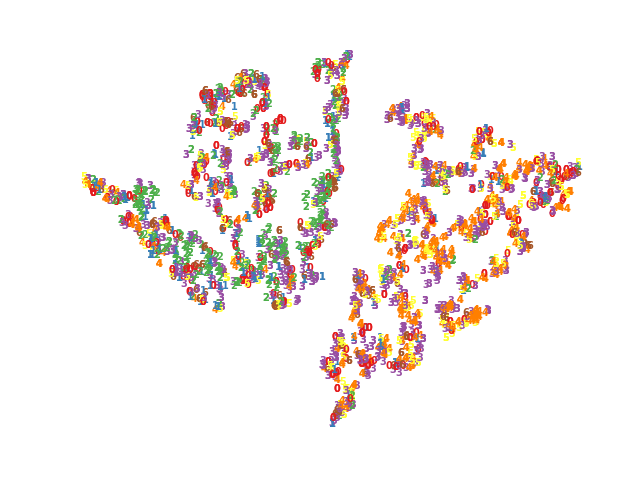}}
\vspace{3pt}
\centerline{\includegraphics[width=\textwidth]{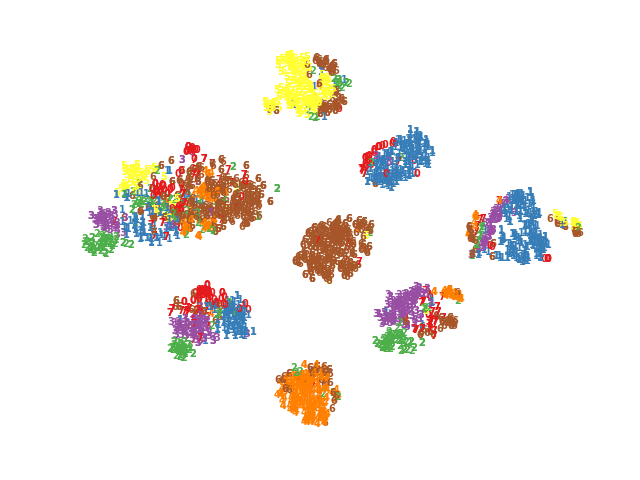}}
\vspace{3pt}
\centerline{(b) SDCN}
\end{minipage}
\begin{minipage}{0.139\linewidth}
\centerline{\includegraphics[width=\textwidth]{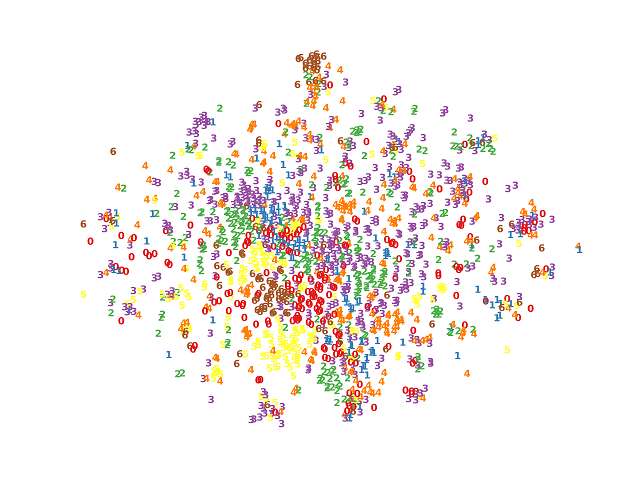}}
\vspace{3pt}
\centerline{\includegraphics[width=\textwidth]{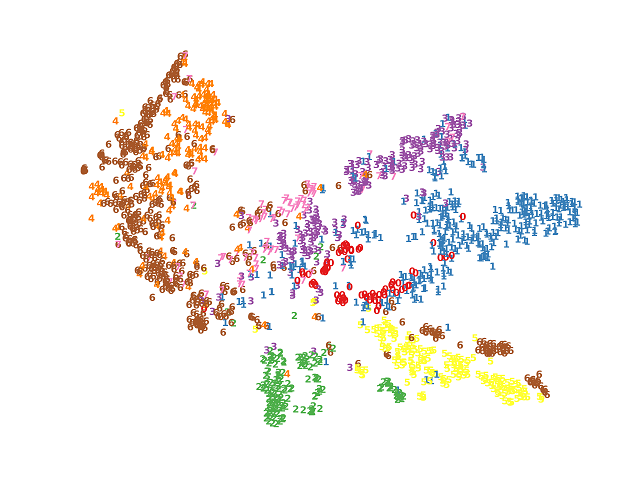}}
\vspace{3pt}
\centerline{(c) AutoSSL}
\end{minipage}
\begin{minipage}{0.139\linewidth}
\centerline{\includegraphics[width=\textwidth]{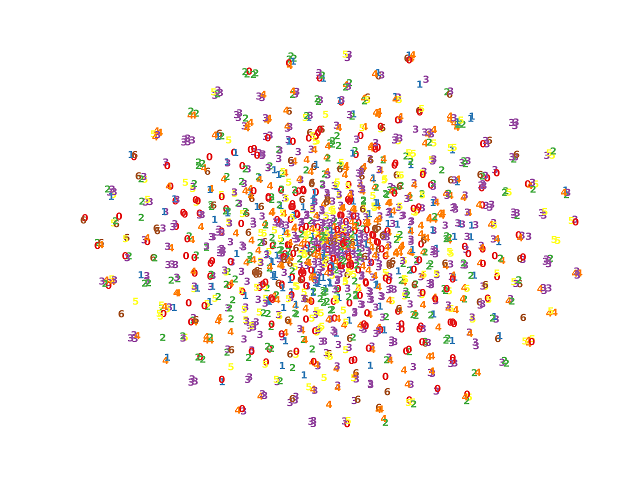}}
\vspace{3pt}
\centerline{\includegraphics[width=\textwidth]{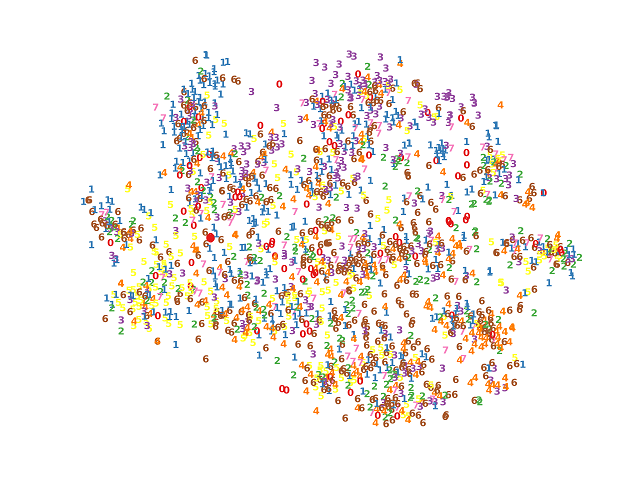}}
\vspace{3pt}
\centerline{(d) AFGRL}
\end{minipage}
\begin{minipage}{0.139\linewidth}
\centerline{\includegraphics[width=\textwidth]{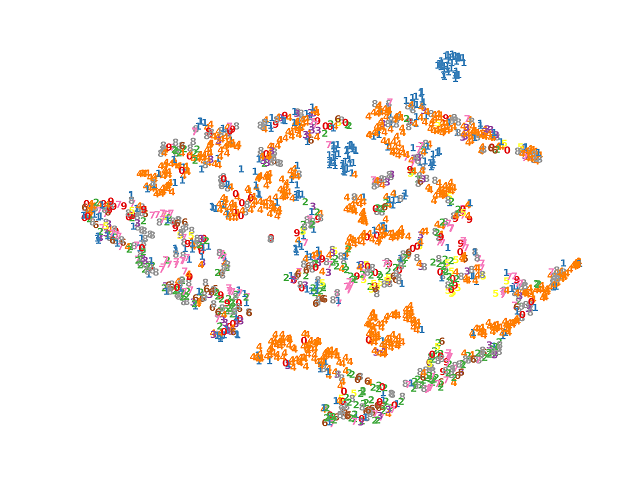}}
\vspace{3pt}
\centerline{\includegraphics[width=\textwidth]{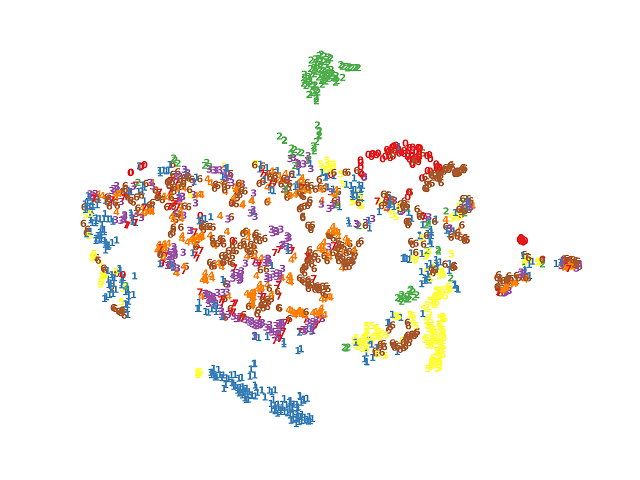}}
\vspace{3pt}
\centerline{(e) GDCL}
\end{minipage}
\begin{minipage}{0.139\linewidth}
\centerline{\includegraphics[width=\textwidth]{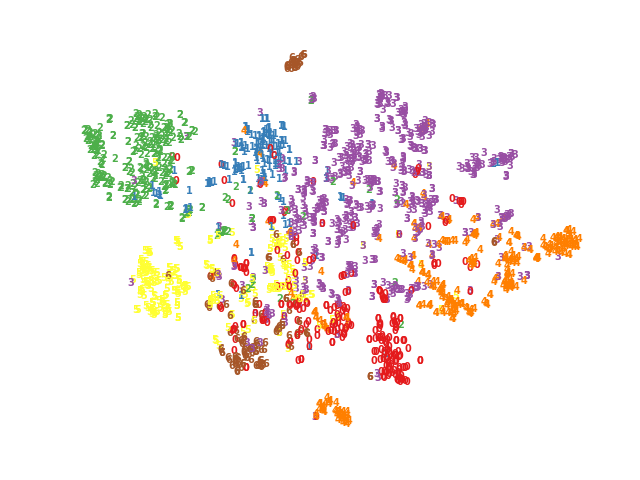}}
\vspace{3pt}
\centerline{\includegraphics[width=\textwidth]{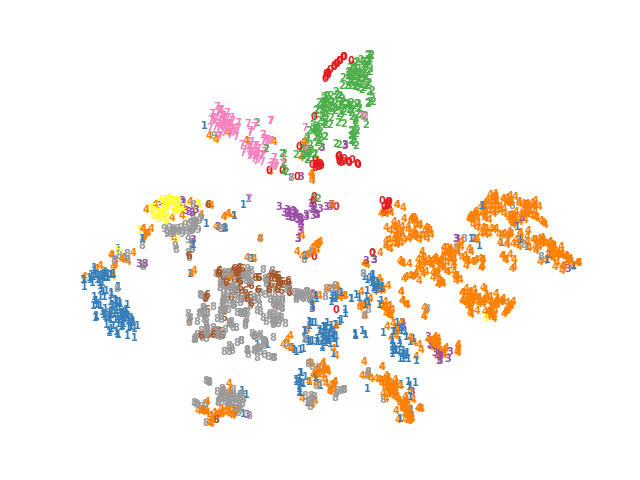}}
\vspace{3pt}
\centerline{(f) ProGCL}
\end{minipage}
\begin{minipage}{0.139\linewidth}
\centerline{\includegraphics[width=\textwidth]{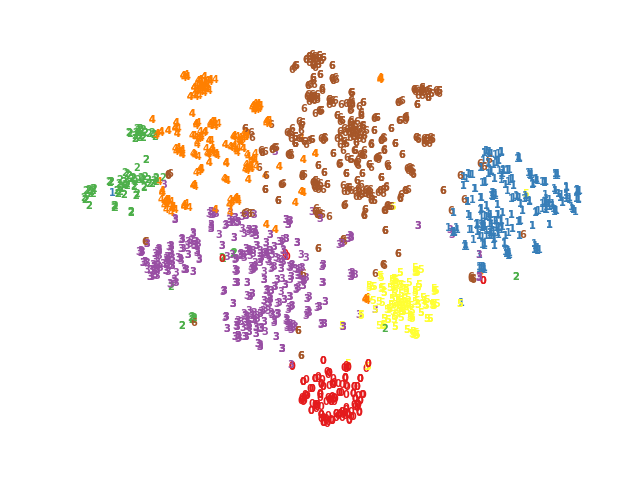}}
\vspace{3pt}
\centerline{\includegraphics[width=\textwidth]{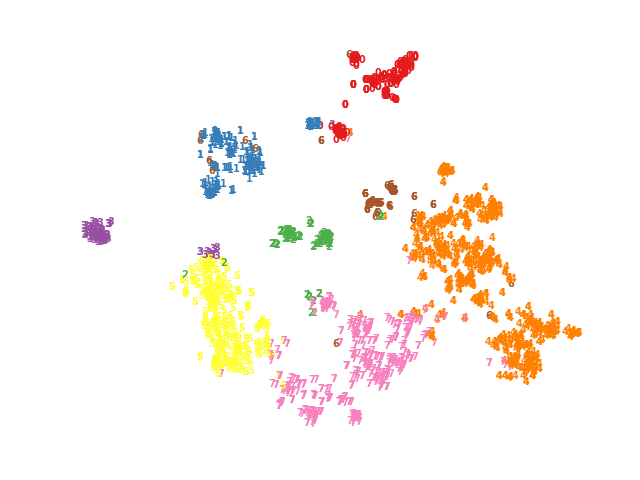}}
\vspace{3pt}
\centerline{(g) Ours}
\end{minipage}
\caption{2D $t$-SNE visualization of seven methods on two benchmark datasets. The first row and second row corresponds to CORA and AMAP dataset, respectively.}
\label{t_SNE}  
\end{figure*}

\subsection{Experimental Setup}
All experimental results are obtained from the desktop computer with the Intel Core i7-7820x CPU, one NVIDIA GeForce RTX 2080Ti GPU, 64GB RAM, and the PyTorch deep learning platform. The training epoch number is set to 400 and we conduct ten runs for all methods. For the baselines, we adopt their source with original settings and reproduce the results. To our model, both the attribute encoders and structure encoders are two parameters un-shared one-layer MLPs with 500 hidden units for UAT/AMAP and 1500 hidden units for other datasets. The learnable trade-off $\alpha$ is set to $0.99999$ as initialization and reduces to around $0.4$ in our experiments as shown in Figure 4 of Appendix. The hyper-parameter settings are summarized in Table 1 of Appendix. The clustering performance is evaluated by four metrics, i.e., ACC, NMI, ARI, and F1, which are widely used in both deep clustering \cite{DCRN,xiawei_3,SDCN,DFCN,IDCRN} and traditional clustering \cite{ZHOU_1,tiejian_1,tiejian_2,suyuan_1,mansheng_1,mansheng_2,liliang_1,liliang_2,zhangpei_1,zhangpei_2,wan_1,siwei_1,siwei_2}.


\subsection{Performance Comparison}
To demonstrate the superiority of our proposed HSAN, we conduct extensive comparison experiments on six benchmark datasets. Concretely, we category these thirteen state-of-the-art deep graph clustering methods into three types, i.e., classical deep graph clustering methods \cite{MGAE,DAEGC,SDCN,DFCN}, contrastive deep graph clustering methods \cite{AGE,MVGRL,AGC-DRR,DCRN,AFGRL,AutoSSL}, and hard sample mining methods \cite{GDCL,ProGCL}. From the results in Table \ref{COMPARE_TABLE}, we have three conclusions as follows. 1) Firstly, compared with the classical deep graph clustering methods, our method achieves promising performance since the contrastive mechanism helps the network capture more potential supervision information. 2) Besides, our proposed HSAN can surpass other contrastive methods thanks to our hard sample mining strategy, which guides the network to focus more the hard samples. 3) Furthermore, the existing hard sample mining methods overlook the hard positive samples and the structural information in the hardness measurement, thus limiting the discriminative capability. Take the CORA dataset as an example, our method surpasses ProGCL \cite{ProGCL} 18.99 \% with the NMI metric. In summary, these experimental results verify the superiority of our proposed methods. Moreover, due to the limitations of paper pages, additional comparison experimental results of nine baselines can be found in Table 2 of the Appendix. The corresponding results also demonstrate the superiority of our proposed HSAN.

\subsection{Ablation Study}
\begin{figure}[h]
\centering
\small
\begin{minipage}{0.49\linewidth}
\centerline{\includegraphics[width=0.9\textwidth]{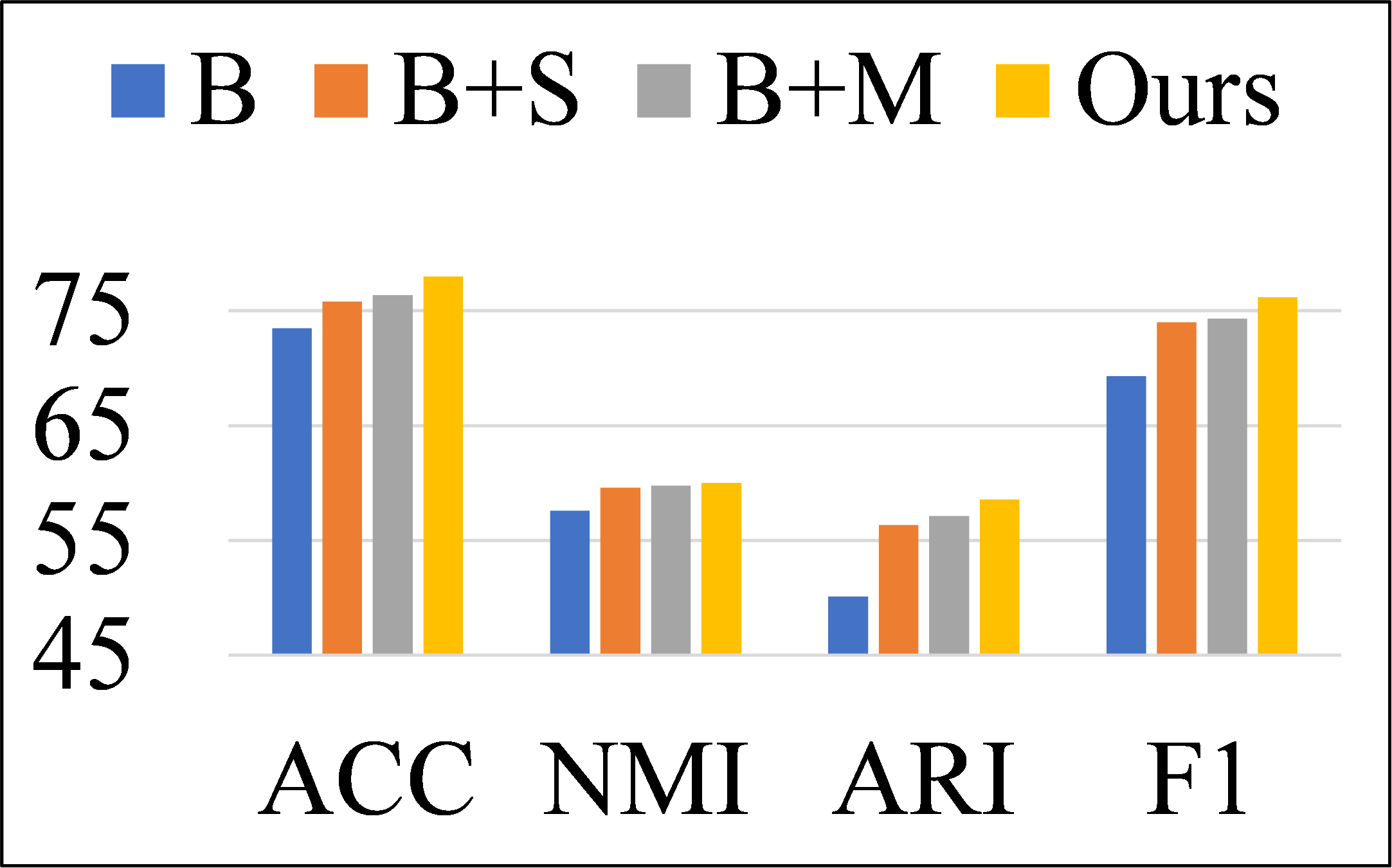}}
\centerline{(a) CORA}
\vspace{3pt}
\centerline{\includegraphics[width=0.9\textwidth]{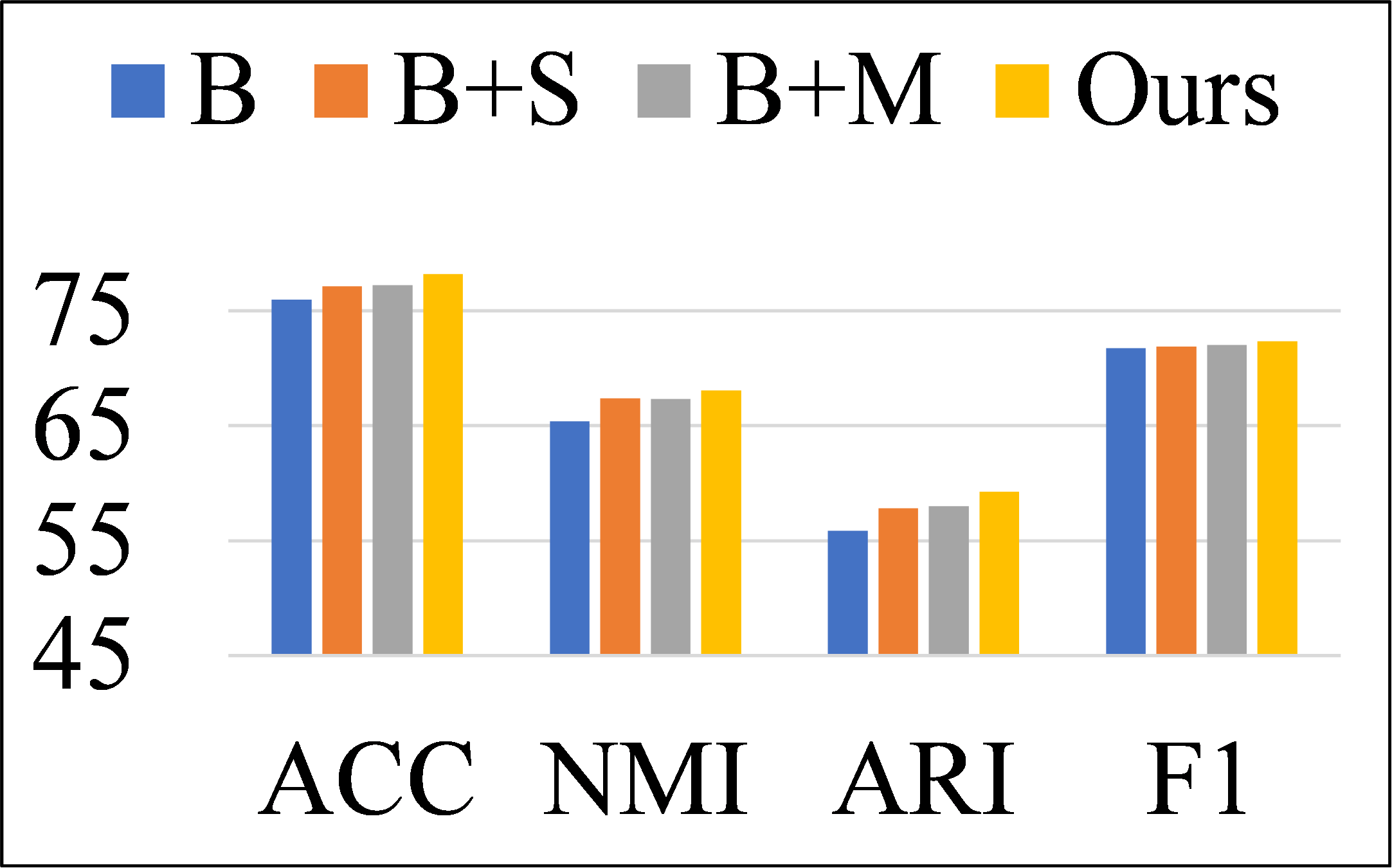}}
\centerline{(b) AMAP}
\vspace{3pt}
\centerline{\includegraphics[width=0.9\textwidth]{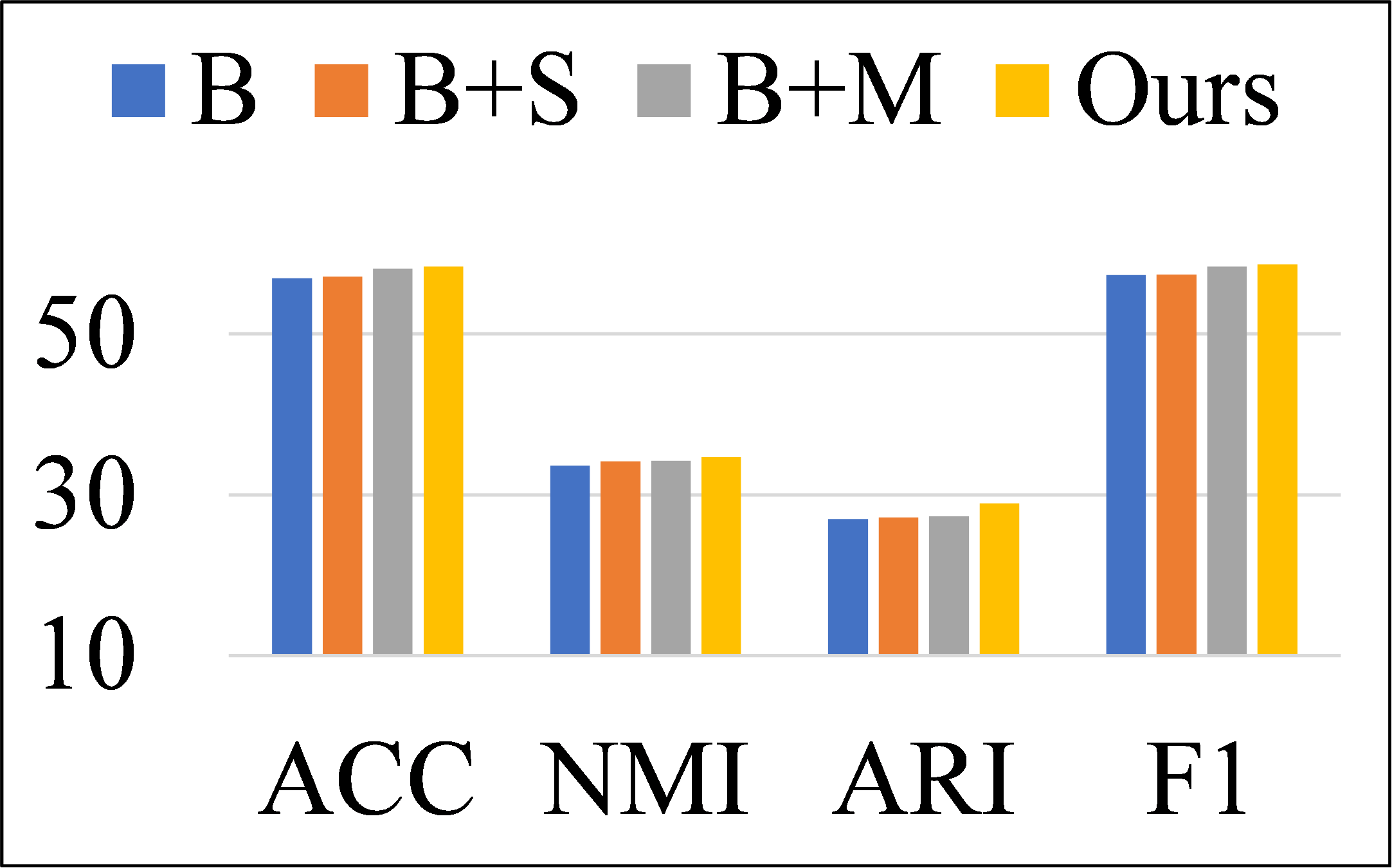}}
\centerline{(c) EAT}
\end{minipage}
\begin{minipage}{0.49\linewidth}
\centerline{\includegraphics[width=0.9\textwidth]{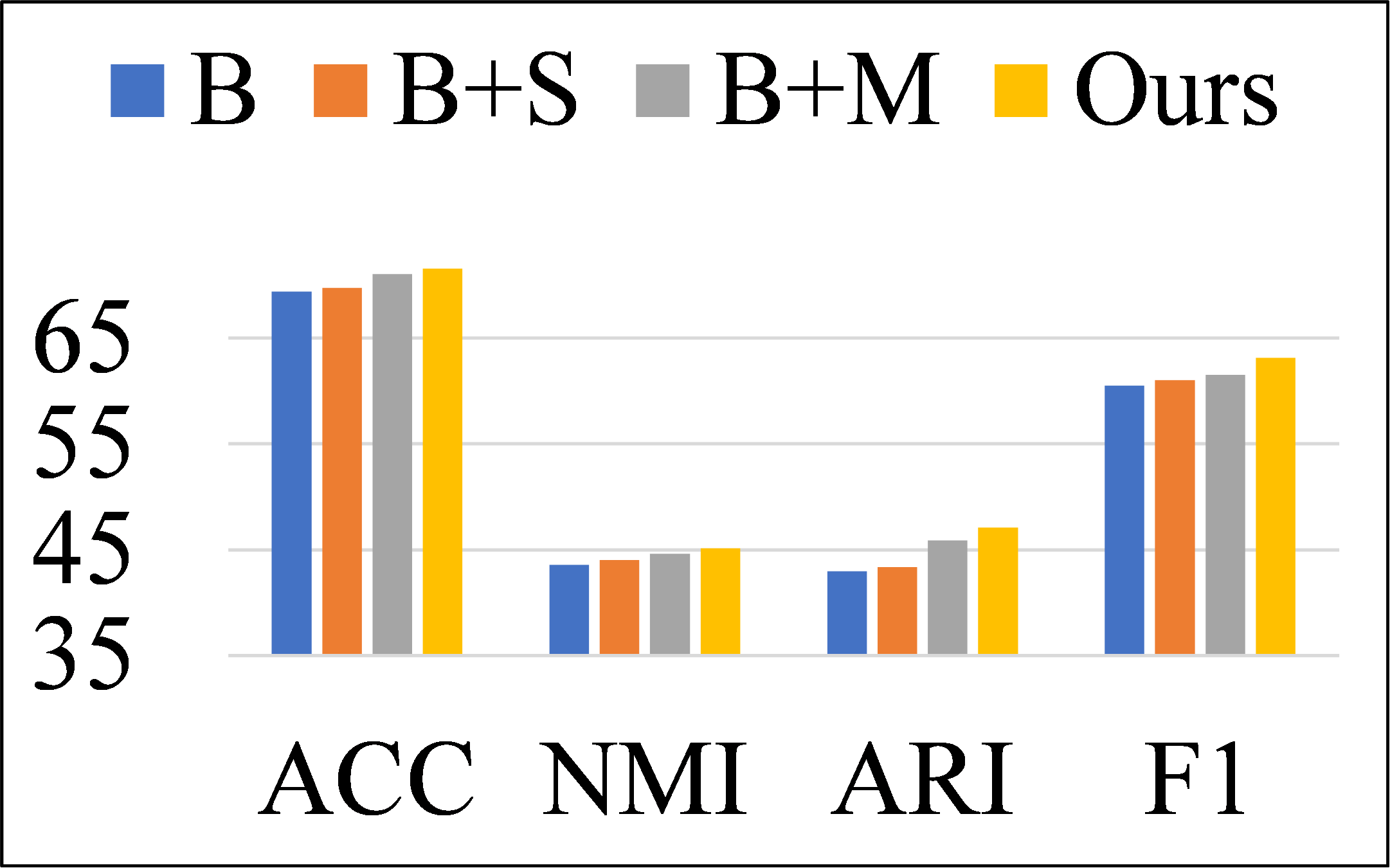}}
\centerline{(d) CITE}
\vspace{3pt}
\centerline{\includegraphics[width=0.9\textwidth]{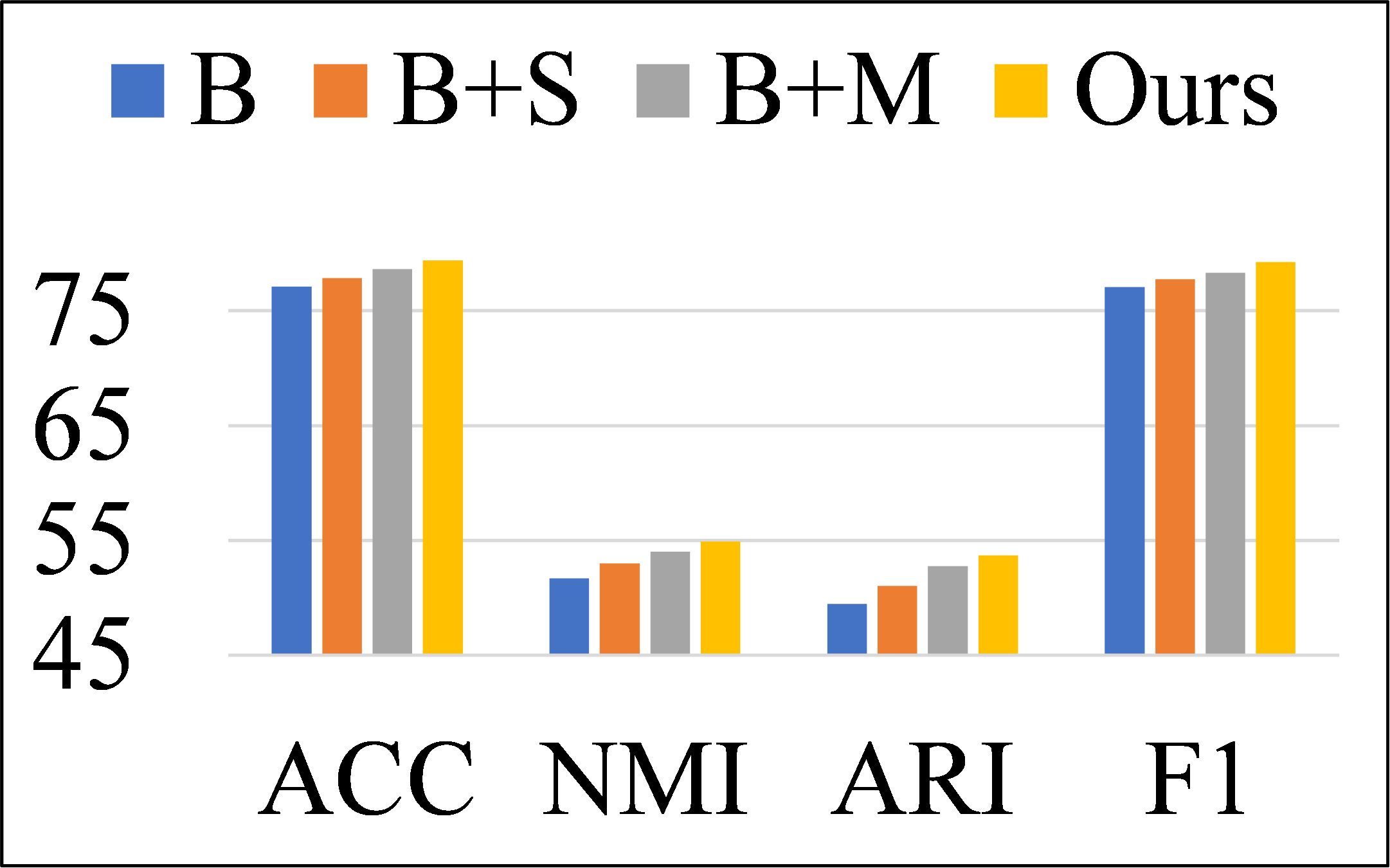}}
\centerline{(e) BAT}
\vspace{3pt}
\centerline{\includegraphics[width=0.9\textwidth]{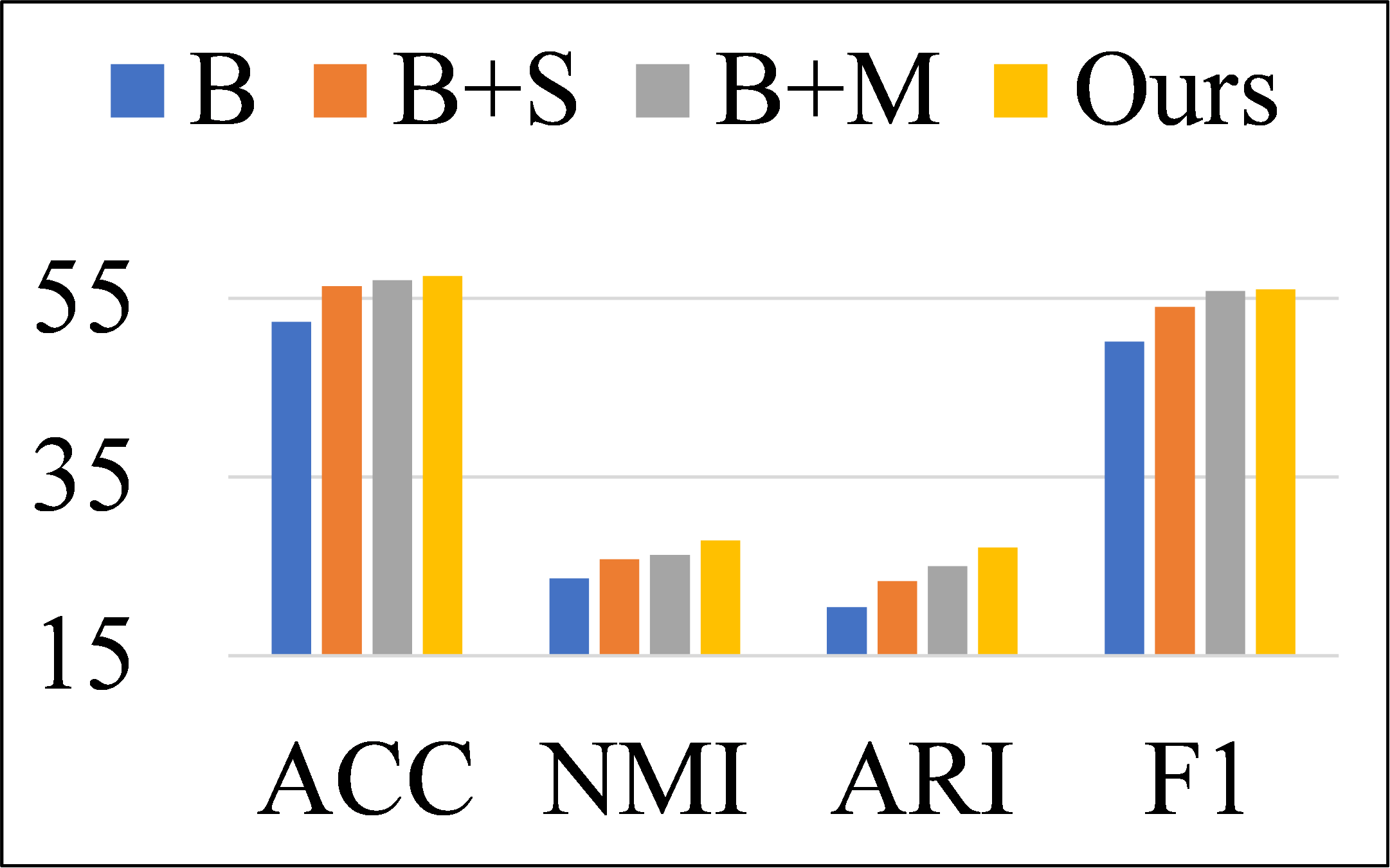}}
\centerline{(f) UAT}
\end{minipage}
\caption{Ablation studies of the proposed similarity function $\mathcal{S}$ and weight modulating function $\mathcal{M}$ on six datasets.}
\label{ablation_figure}
\end{figure}

In this section, we conduct ablation studies to verify the effectiveness of the proposed attribute-structure similarity function $\mathcal{S}$ and weight modulating function $\mathcal{M}$. Concretely, in Figure \ref{ablation_figure}, we denote ``B'' as the baseline. In addition, ``B+S'', ``B+M'', and ``Ours'' denotes the baseline with $\mathcal{S}$, $\mathcal{M}$, and both, respectively. From the results in Figure \ref{ablation_figure}, we have three observations as follows. 1) Our proposed attribute-structure similarity function $\mathcal{S}$ improves the performance of the baseline. The reason is that $\mathcal{S}$ measures the similarity between samples by considering both attribute and structure information, thus better revealing the potential relation between samples. 2) ``B+M'' can improve the performance of ``B''. It indicates that the proposed weight modulating function $\mathcal{M}$ enhances the discriminative capability of samples by guiding our network to focus on the hard sample pairs. 3) The combination of $\mathcal{S}$ and $\mathcal{M}$ achieves the most superior clustering performance. For example, ``Ours'' exceeds ``B'' 8.48 \% with the NMI metric on the CORA dataset. Overall, the effectiveness of our proposed attribute-structure similarity function $\mathcal{S}$ and weight modulating function $\mathcal{M}$ is verified by extensive experimental results.

\subsection{Analysis}
\subsubsection{Hyper-parameter Analysis}
\begin{figure}[!t]
\centering
\small
\begin{minipage}{0.325\linewidth}
\centerline{\includegraphics[width=1\textwidth]{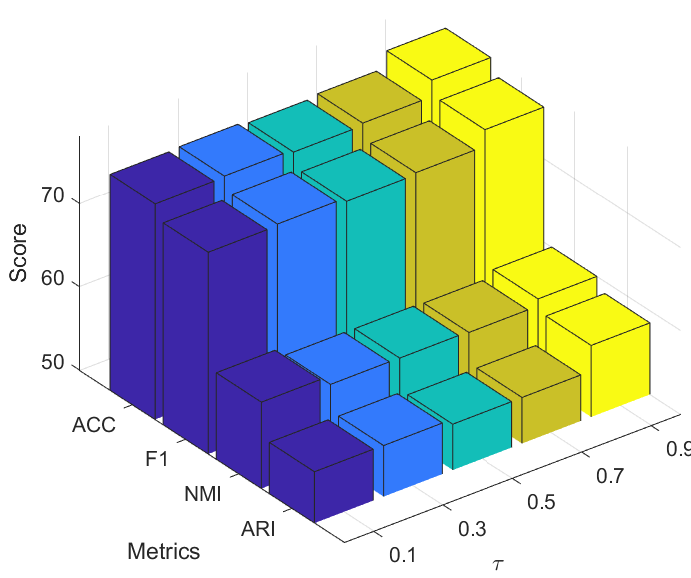}}
\vspace{3pt}
\centerline{(a) CORA}
\centerline{\includegraphics[width=\textwidth]{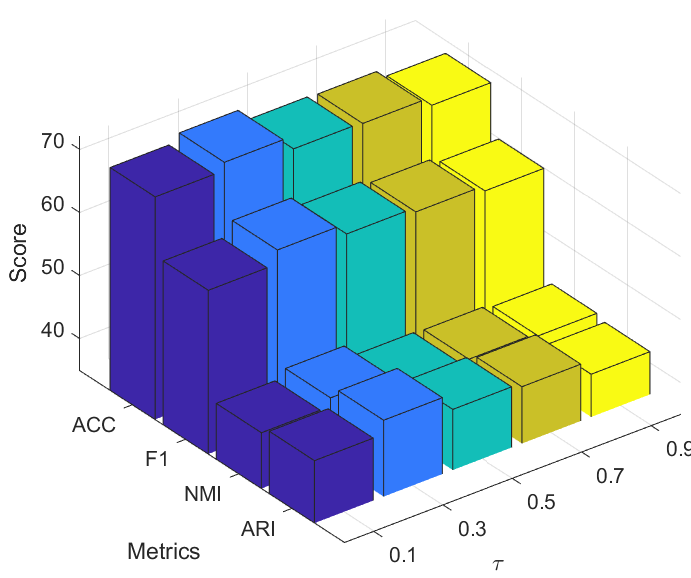}}
\vspace{3pt}
\centerline{(b) CITE}
\end{minipage}
\begin{minipage}{0.325\linewidth}
\centerline{\includegraphics[width=\textwidth]{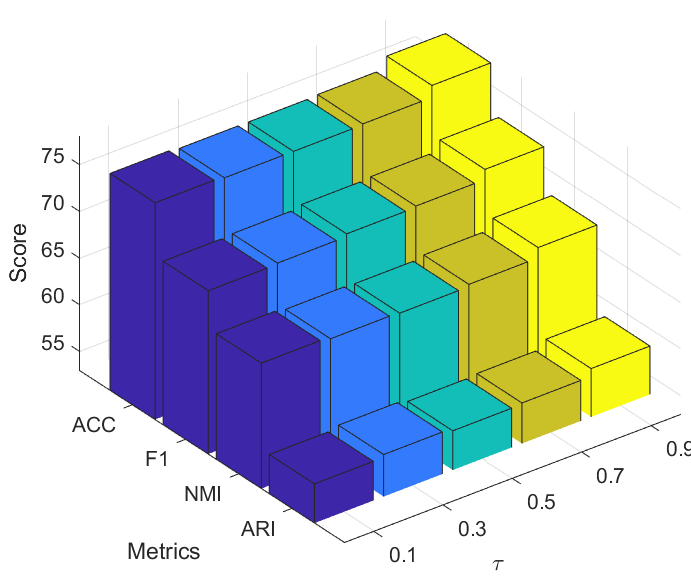}}
\vspace{3pt}
\centerline{(c) AMAP}
\centerline{\includegraphics[width=1\textwidth]{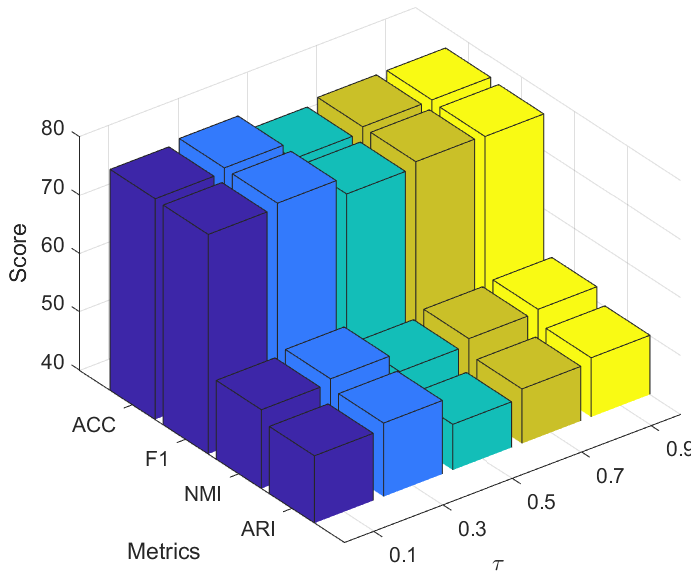}}
\vspace{3pt}
\centerline{(d) BAT}
\end{minipage}
\begin{minipage}{0.325\linewidth}
\centerline{\includegraphics[width=\textwidth]{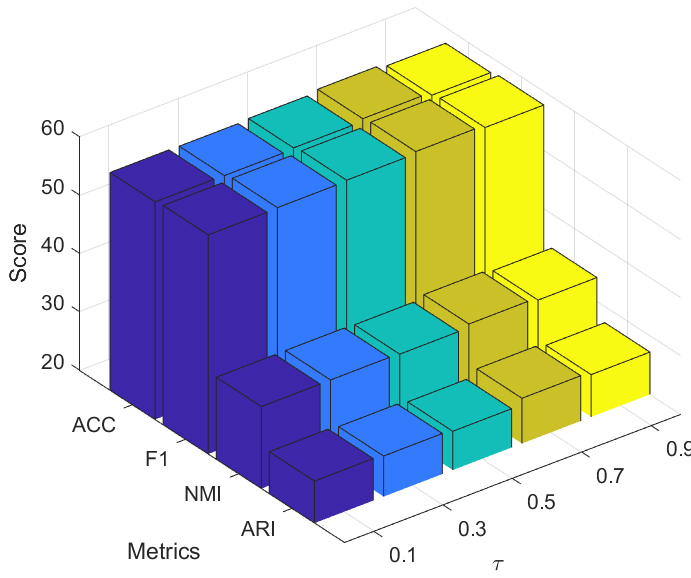}}
\vspace{3pt}
\centerline{(e) EAT}
\vspace{3pt}
\centerline{\includegraphics[width=\textwidth]{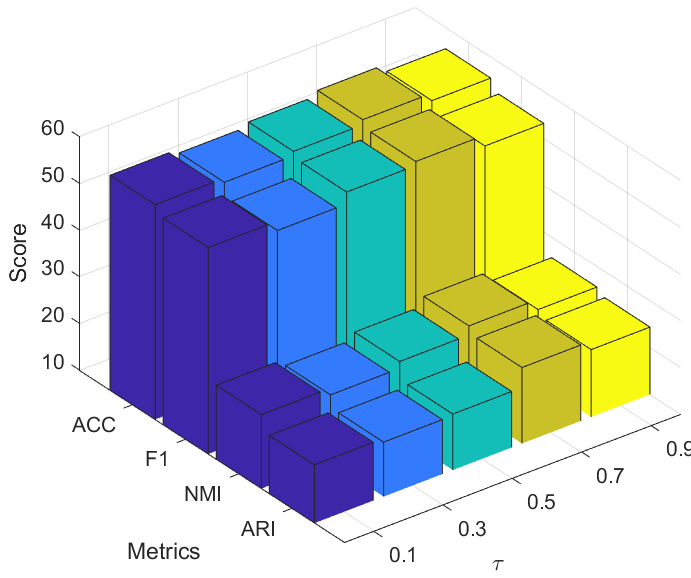}}
\vspace{3pt}
\centerline{(f) UAT}
\end{minipage}
\caption{Analysis of the confidence hyper-parameter $\tau$.}
\label{tau}
\end{figure}

In this section, we analyze the hyper-parameters $\tau$ and $\beta$ in our method. For the confidence $\tau$, we select it in $\{0.1, 0.3, 0.5, 0.7, 0.9\}$. As shown in Figure \ref{tau}, we observe that our method achieves promising performance when $\tau \in [0.1, 0.3]$ on BAT / CITE datasets and when $\tau \in [0.7, 0.9]$ on other datasets. In this paper, the confidence is set to a fixed value, thus a possible future work is to design a learnable or dynamical confidence. Besides, we analyze $\beta$ in Appendix. Two conclusions are deduced as follows. 1) The focusing factor $\beta$ controls the down-weighing rate of easy sample pairs. When $\beta$ increases, the down-weighting rate of easy sample pairs also increases and vice versa. 2) HSAN is not sensitive to $\beta$. Experimental evidence can be found in Figure 1-3 in Appendix.


\subsubsection{Visualization Analysis}
To further demonstrate the superiority of HSAN intuitively, we conduct 2D $t$-SNE \cite{T_SNE} on the learned node embeddings in Figure \ref{t_SNE}. It is observed that our HSAN can better reveal the cluster structure compared with other baselines.

\subsubsection{Convergence Analysis}
\begin{figure}[!t]
\centering
\small
\begin{minipage}{0.49\linewidth}
\centerline{\includegraphics[width=0.9\textwidth]{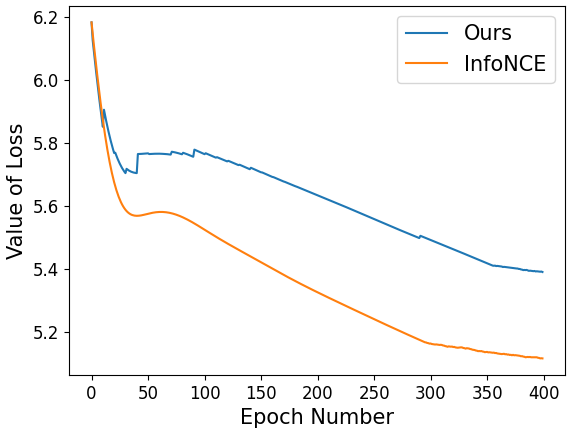}}
\centerline{(a) Loss on BAT}
\end{minipage}
\begin{minipage}{0.49\linewidth}
\centerline{\includegraphics[width=0.9\textwidth]{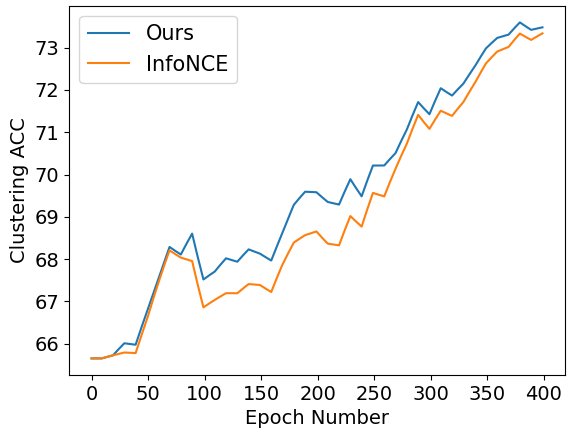}}
\centerline{(b) ACC on BAT}
\end{minipage}
\caption{Convergence analysis on BAT dataset.}
\label{convergence}
\end{figure}

In addition, we analyze the convergence of our proposed loss. Specifically, we plot the trend of loss and the clustering ACC of our method in Figure \ref{convergence}. Here, ``Ours'' and ``infoNCE'' denotes our method with our hard sample aware contrastive loss, and the infoNCE loss, respectively. We observe 1) the loss and ACC gradually converge after 350 epochs. 2) after 50 epochs, ``Ours'' calculates the larger loss value while achieving better performance compared with ``infoNCE''. The reason is that our loss down-weights the easy sample pairs while up-weighting the hard sample pairs, thus increasing the loss value. Meanwhile, the proposed loss guides the network to focus on the hard sample pairs, leading to better performance.

\section{Conclusion}
In this paper, we propose a Hard Sample Aware Network (HSAN) to mine the hard samples in contrastive deep graph clustering. Concretely, we first design the attribute and structure encoders to embed the attribute and structure of samples into the latent space. Then a comprehensive similarity measure criterion is proposed to calculate the sample similarities by considering both attribute and structure information, thus better revealing the potential sample relations. Furthermore, guided by the high-confidence clustering information, we propose a general dynamic weight modulating function to up-weight the hard sample pairs while down-weighting the negative ones. In this manner, the proposed hard sample aware contrastive loss forces the network to focus on both positive and negative sample pairs, thus further improving the discriminative capability. The time and space analysis of the proposed loss demonstrate that it will not bring large time or space costs compared with the classical infoNCE loss. Experiments demonstrate the effectiveness and superiority of our proposed method. In this work, the confidence parameter is set to a fixed value, thus one future work is to design a learnable or adaptive confidence parameter.

\newpage
\section{Acknowledgments}
This work was supported by the National Key R\&D Program of China (project no. 2020AAA0107100) and the National Natural Science Foundation of China (project no. 61922088, 61976196, 62006237, and 61872371).




\bibliography{aaai23}

\end{document}


\maketitle

\appendix

\section{Details of the Proposed Method}
In this section, we introduce the detailed implementation of our method with PyTorch-style pseudo codes in Algorithm \ref{code}.

\begin{algorithm}[h]
	\caption{PyTorch-style Pseudo Code of Our Method.}
	\label{code}
	\definecolor{codeblue}{rgb}{0.25,0.5,0.5}
	\lstset{
		backgroundcolor=\color{white},
		basicstyle=\fontsize{7.2pt}{7.2pt}\ttfamily\selectfont,
		columns=fullflexible,
		breaklines=true,
		captionpos=b,
		commentstyle=\fontsize{7.2pt}{7.2pt}\color{codeblue},
		keywordstyle=\fontsize{7.2pt}{7.2pt},
	}
    
    \begin{lstlisting}[language=python]
    # X_: Filtered Attribute, A: Original Structure
    # AE: Attribute Encoders, SE: Structure Encoders
    # S: Attribute-Structure Similarity Function
    # M: Weight Modulating Function
    for epoch in range(epoch_num):
        # Attribute and Structure Encoding
        Z1 = F.normalization(AE1(X_),dim=1,p=2)
        Z2 = F.normalization(AE2(X_),dim=1,p=2)
        E1 = F.normalization(SE1(A),dim=1,p=2)
        E2 = F.normalization(SE2(A),dim=1,p=2)
        
        # Attribute-Structure Similarity
        S = alpha * Z1.T @ Z2 + (1-alpha) * E1.T @ E2
        
        # Clustering and Pseudo Label Generation
        clu_res, P, H = clustering((Z1+Z2/2))
        Q = (P==P.T).int()
        
        # Hard Sample Aware Contrastive Learning
        M = abs(Q-min_max_norm(S))**beta
        w[H, :][:, H] = M[H, :][:, H]
        loss = (pos*w / (pos*w+neg*w)).mean()
        
        # optimization
        loss.backward()
        optimizer.step()
    
    clu_res, _, _ = clustering((Z1+Z2/2))
    return clu_res
    
    
    
    \end{lstlisting}
    
\end{algorithm}

\section{Additional Experiments}
In this section, due to the limitation of the original paper pages, we conduct additional experiments including comparison experiments and visualization analysis experiments.

\subsection{Statistics and Hyper-parameter Settings}
To guarantee reproducibility, we report the statistics summary and hyper-parameter settings of our proposed method in Table \ref{DATASET_INFO}.

\subsection{Additional Comparison Experiments}
Due to the limitation of the paper pages, we report additional comparison experimental results in this section. Concretely, we compare our proposed method with three categories of the existing methods, i.e., deep clustering method including DCN \cite{AE_K_MEANS}, DEC \cite{DEC}, IDEC \cite{IDEC}, AdaGAE \cite{AdaGAE}, classical deep graph clustering method including ARVGA \cite{ARGA_conf}, $\text{SDCN}_Q$ \cite{SDCN}, and contrastive deep graph clustering method including GCA \cite{GCA}, MCGC \cite{MCGC}, SUBLIME \cite{SUBLIME}. From the results in Table \ref{compare}, we have three conclusions as follows. 1) Deep clustering methods achieve un-promising performance since they overlook the graph structure. 2) Classical deep graph clustering methods can not compare with ours. The reason is that contrastive learning enhances the supervision information capture capability of our method. 3) Our proposed method can achieve better performance than the contrastive method thanks to the hard sample mining, which improves the discriminative capability of our network.

\begin{table*}[!t]
\centering
\small
\caption{Statistics and hyper-parameter settings of six benchmark datasets.}
\scalebox{1.2}{
\begin{tabular}{c|cclclclclclcl}
\hline
                                           & \textbf{Dataset}       & \multicolumn{2}{c}{\textbf{CORA}} & \multicolumn{2}{c}{\textbf{CITE}} & \multicolumn{2}{c}{\textbf{AMAP}} & \multicolumn{2}{c}{\textbf{BAT}} & \multicolumn{2}{c}{\textbf{EAT}} & \multicolumn{2}{c}{\textbf{UAT}} \\ \hline
\multirow{5}{*}{\textbf{Statistics}}       & Type                   & \multicolumn{2}{c}{Graph}         & \multicolumn{2}{c}{Graph}         & \multicolumn{2}{c}{Graph}         & \multicolumn{2}{c}{Graph}        & \multicolumn{2}{c}{Graph}        & \multicolumn{2}{c}{Graph}        \\
                                           & \# Samples             & \multicolumn{2}{c}{2708}          & \multicolumn{2}{c}{3327}          & \multicolumn{2}{c}{7650}          & \multicolumn{2}{c}{131}          & \multicolumn{2}{c}{399}          & \multicolumn{2}{c}{1190}         \\
                                           & \# Dimensions          & \multicolumn{2}{c}{1433}          & \multicolumn{2}{c}{3703}          & \multicolumn{2}{c}{745}           & \multicolumn{2}{c}{81}           & \multicolumn{2}{c}{203}          & \multicolumn{2}{c}{239}          \\
                                           & \# Edges               & \multicolumn{2}{c}{5429}          & \multicolumn{2}{c}{4732}          & \multicolumn{2}{c}{119081}        & \multicolumn{2}{c}{1038}         & \multicolumn{2}{c}{5994}         & \multicolumn{2}{c}{13599}        \\
                                           & \# Classes             & \multicolumn{2}{c}{7}             & \multicolumn{2}{c}{6}             & \multicolumn{2}{c}{8}             & \multicolumn{2}{c}{4}            & \multicolumn{2}{c}{4}            & \multicolumn{2}{c}{4}            \\ \hline
\multirow{4}{*}{\makecell[c]{\textbf{Hyper-} \\ \textbf{parameters}}} & $\tau$          & \multicolumn{2}{c}{0.9}           & \multicolumn{2}{c}{0.3}           & \multicolumn{2}{c}{0.9}           & \multicolumn{2}{c}{0.3}          & \multicolumn{2}{c}{0.7}          & \multicolumn{2}{c}{0.7}          \\
                                           & $\beta$                   & \multicolumn{2}{c}{1}             & \multicolumn{2}{c}{2}             & \multicolumn{2}{c}{3}             & \multicolumn{2}{c}{5}            & \multicolumn{2}{c}{5}            & \multicolumn{2}{c}{5}            \\
                                           & $t$                      & \multicolumn{2}{c}{2}             & \multicolumn{2}{c}{2}             & \multicolumn{2}{c}{3}             & \multicolumn{2}{c}{6}            & \multicolumn{2}{c}{6}            & \multicolumn{2}{c}{6}            \\
                                           & Learning rate & \multicolumn{2}{c}{$10^{-3}$}            & \multicolumn{2}{c}{$10^{-3}$}            & \multicolumn{2}{c}{$10^{-5}$}            & \multicolumn{2}{c}{$10^{-3}$}           & \multicolumn{2}{c}{$10^{-4}$}           & \multicolumn{2}{c}{$10^{-4}$}           \\ \hline
\end{tabular}}
\label{DATASET_INFO}
\end{table*}

\begin{table*}[]
\centering
\caption{Additional comparison experiments on six benchmark datasets. The clustering performance is evaluated by four metrics with mean value and standard deviation. The {\color[HTML]{FF0000}red} values indicate the best results.}
\scalebox{0.75}{
\begin{tabular}{c|c|cccc|cc|ccc|c}
\hline
{\color[HTML]{000000} }                                   &                                   & \multicolumn{4}{c|}{\textbf{Deep Clustering}}                                         & \multicolumn{2}{c|}{\makecell[c]{\textbf{Classical Deep} \\ \textbf{Graph Clustering}}} & \multicolumn{3}{c|}{\textbf{Contrastive Deep Graph Clustering}}                        & \textbf{}                                  \\ \cline{3-12} 
{\color[HTML]{000000} }                                   &                                   & \textbf{DCN}        & \textbf{DEC}        & \textbf{IDEC}       & \textbf{AdaGAE}     & \textbf{ARGVA}                & \textbf{SDCN\_Q}              & \textbf{GCA}        & \textbf{MCGC}       & \textbf{SUBLIME}                           & \textbf{xxx}                               \\
\multirow{-3}{*}{{\color[HTML]{000000} \textbf{Dataset}}} & \multirow{-3}{*}{\textbf{Metric}} & \textbf{ICML 17}    & \textbf{ICML 16}    & \textbf{IJCAI 17}   & \textbf{TPAMI 21}   & \textbf{IJCAI 19}             & \textbf{WWW 20}               & \textbf{WWW 21}     & \textbf{NeurIPS 21} & \textbf{WWW 22}                            & \textbf{Ours}                              \\ \hline
                                                          & \textbf{ACC}                      & \textbf{49.38±0.91} & \textbf{46.50±0.26} & \textbf{51.61±1.02} & \textbf{50.06±1.58} & \textbf{65.97±1.15}           & \textbf{39.05±3.50}           & \textbf{53.62±0.73} & \textbf{42.85±1.13} & \textbf{71.14±0.74}                        & {\color[HTML]{FF0000} \textbf{77.97±1.56}} \\
                                                          & \textbf{NMI}                      & \textbf{25.65±0.65} & \textbf{23.54±0.34} & \textbf{26.31±1.22} & \textbf{32.19±1.34} & \textbf{49.30±0.54}           & \textbf{17.50±4.88}           & \textbf{46.87±0.65} & \textbf{24.11±1.00} & \textbf{53.88±1.02}                        & {\color[HTML]{FF0000} \textbf{60.01±1.03}} \\
                                                          & \textbf{ARI}                      & \textbf{21.63±0.58} & \textbf{15.13±0.42} & \textbf{22.07±1.53} & \textbf{28.25±0.98} & \textbf{41.28±1.93}           & \textbf{11.31±5.37}           & \textbf{30.32±0.98} & \textbf{14.33±1.26} & {\color[HTML]{330001} \textbf{50.15±0.14}} & {\color[HTML]{FF0000} \textbf{58.58±2.70}} \\
\multirow{-4}{*}{\textbf{CORA}}                           & \textbf{F1}                       & \textbf{43.71±1.05} & \textbf{39.23±0.17} & \textbf{47.17±1.12} & \textbf{53.53±1.24} & \textbf{63.71±2.09}           & \textbf{24.71±4.85}           & \textbf{45.73±0.47} & \textbf{35.16±0.91} & \textbf{63.11±0.58}                        & {\color[HTML]{FF0000} \textbf{76.17±1.40}} \\ \hline
                                                          & \textbf{ACC}                      & \textbf{57.08±0.13} & \textbf{55.89±0.20} & \textbf{60.49±1.42} & \textbf{54.01±1.11} & \textbf{59.31±1.38}           & \textbf{61.67±1.05}           & \textbf{60.45±1.03} & \textbf{64.76±0.07} & \textbf{68.25±1.21}                        & {\color[HTML]{FF0000} \textbf{71.55±0.80}} \\
                                                          & \textbf{NMI}                      & \textbf{27.64±0.08} & \textbf{28.34±0.30} & \textbf{27.17±2.40} & \textbf{27.79±0.47} & \textbf{31.80±0.81}           & \textbf{34.39±1.22}           & \textbf{36.15±0.78} & \textbf{39.11±0.06} & \textbf{43.15±0.14}                        & {\color[HTML]{FF0000} \textbf{45.16±0.74}} \\
                                                          & \textbf{ARI}                      & \textbf{29.31±0.14} & \textbf{28.12±0.36} & \textbf{25.70±2.65} & \textbf{24.19±0.85} & \textbf{31.28±1.22}           & \textbf{35.50±1.49}           & \textbf{35.20±0.96} & \textbf{37.54±0.12} & \textbf{44.21±0.54}                        & {\color[HTML]{FF0000} \textbf{47.10±1.12}} \\
\multirow{-4}{*}{\textbf{CITE}}                       & \textbf{F1}                       & \textbf{53.80±0.11} & \textbf{52.62±0.17} & \textbf{61.62±1.39} & \textbf{51.11±0.64} & \textbf{56.05±1.13}           & \textbf{57.82±0.98}           & \textbf{56.42±0.94} & \textbf{59.64±0.05} & \textbf{63.12±0.42}                        & {\color[HTML]{FF0000} \textbf{63.13±1.79}} \\ \hline
                                                          & \textbf{ACC}                      & \textbf{48.25±0.08} & \textbf{47.22±0.08} & \textbf{47.62±0.08} & \textbf{67.70±0.54} & \textbf{61.46±2.71}           & \textbf{35.53±0.39}           & \textbf{56.81±1.44} & \textbf{71.64±0.52} & \textbf{27.22±1.56}                        & {\color[HTML]{FF0000} \textbf{77.40±0.33}} \\
                                                          & \textbf{NMI}                      & \textbf{38.76±0.30} & \textbf{37.35±0.05} & \textbf{37.83±0.08} & \textbf{55.96±0.87} & \textbf{53.25±1.91}           & \textbf{27.90±0.40}           & \textbf{48.38±2.38} & \textbf{61.54±0.36} & \textbf{06.37±1.89}                        & {\color[HTML]{FF0000} \textbf{67.44±0.33}} \\
                                                          & \textbf{ARI}                      & \textbf{20.80±0.47} & \textbf{18.59±0.04} & \textbf{19.24±0.07} & \textbf{46.20±0.45} & \textbf{38.44±4.69}           & \textbf{15.27±0.37}           & \textbf{26.85±0.44} & \textbf{43.23±0.64} & \textbf{05.36±2.14}                        & {\color[HTML]{FF0000} \textbf{58.18±0.48}} \\
\multirow{-4}{*}{\textbf{AMAP}}                           & \textbf{F1}                       & \textbf{47.87±0.20} & \textbf{46.71±0.12} & \textbf{47.20±0.11} & \textbf{62.95±0.74} & \textbf{58.50±1.70}           & \textbf{34.25±0.44}           & \textbf{53.59±0.57} & \textbf{68.64±0.67} & \textbf{15.97±1.53}                        & {\color[HTML]{FF0000} \textbf{72.11±0.46}} \\ \hline
                                                          & \textbf{ACC}                      & \textbf{47.79±3.95} & \textbf{42.09±2.21} & \textbf{39.62±0.87} & \textbf{43.51±0.48} & \textbf{63.66±0.78}           & \textbf{47.10±2.53}           & \textbf{54.89±0.34} & \textbf{38.93±0.23} & \textbf{45.04±0.19}                        & {\color[HTML]{FF0000} \textbf{78.55±0.72}} \\
                                                          & \textbf{NMI}                      & \textbf{18.03±7.73} & \textbf{14.10±1.99} & \textbf{12.80±1.74} & \textbf{15.84±0.78} & \textbf{40.79±0.95}           & \textbf{19.26±4.33}           & \textbf{38.88±0.23} & \textbf{23.11±0.56} & \textbf{22.03±0.48}                        & {\color[HTML]{FF0000} \textbf{54.25±0.93}} \\
                                                          & \textbf{ARI}                      & \textbf{13.75±6.05} & \textbf{07.99±1.21} & \textbf{07.85±1.31} & \textbf{07.80±0.41} & \textbf{29.89±1.14}           & \textbf{14.41±3.07}           & \textbf{26.69±2.85} & \textbf{8.41±0.32}  & \textbf{14.45±0.87}                        & {\color[HTML]{FF0000} \textbf{52.60±1.11}} \\
\multirow{-4}{*}{\textbf{BAT}}                            & \textbf{F1}                       & \textbf{46.80±3.44} & \textbf{42.63±2.35} & \textbf{40.11±0.99} & \textbf{43.15±0.77} & \textbf{63.90±0.66}           & \textbf{43.73±1.68}           & \textbf{53.71±0.34} & \textbf{32.92±0.25} & \textbf{44.00±0.62}                        & {\color[HTML]{FF0000} \textbf{78.43±0.76}} \\ \hline
                                                          & \textbf{ACC}                      & \textbf{38.85±2.32} & \textbf{36.47±1.60} & \textbf{35.56±1.34} & \textbf{32.83±1.24} & \textbf{50.35±0.41}           & \textbf{33.21±1.93}           & \textbf{48.51±1.55} & \textbf{32.58±0.29} & \textbf{38.80±0.35}                        & {\color[HTML]{FF0000} \textbf{57.89±0.34}} \\
                                                          & \textbf{NMI}                      & \textbf{06.92±2.80} & \textbf{04.96±1.74} & \textbf{04.63±0.97} & \textbf{04.36±1.87} & \textbf{33.68±1.71}           & \textbf{2.74±1.12}            & \textbf{28.36±1.23} & \textbf{7.04±0.56}  & \textbf{14.96±0.75}                        & {\color[HTML]{FF0000} \textbf{34.55±0.44}} \\
                                                          & \textbf{ARI}                      & \textbf{05.11±2.65} & \textbf{03.60±1.87} & \textbf{03.19±0.76} & \textbf{02.47±0.54} & \textbf{23.28±1.55}           & \textbf{1.79±1.12}            & \textbf{19.61±1.25} & \textbf{1.33±0.14}  & \textbf{10.29±0.88}                        & {\color[HTML]{FF0000} \textbf{27.75±0.59}} \\
\multirow{-4}{*}{\textbf{EAT}}                            & \textbf{F1}                       & \textbf{38.75±2.25} & \textbf{34.84±1.28} & \textbf{35.52±1.50} & \textbf{32.39±0.47} & \textbf{48.64±0.52}           & \textbf{32.58±1.69}           & \textbf{48.22±0.33} & \textbf{27.03±0.16} & \textbf{32.31±0.97}                        & {\color[HTML]{FF0000} \textbf{58.18±0.28}} \\ \hline
                                                          & \textbf{ACC}                      & \textbf{46.82±1.14} & \textbf{45.61±1.84} & \textbf{46.90±0.17} & \textbf{52.10±0.87} & \textbf{51.82±0.83}           & \textbf{49.33±0.74}           & \textbf{39.39±1.46} & \textbf{41.93±0.56} & \textbf{48.74±0.54}                        & {\color[HTML]{FF0000} \textbf{57.14±0.67}} \\
                                                          & \textbf{NMI}                      & \textbf{17.18±1.60} & \textbf{16.63±2.39} & \textbf{17.84±0.35} & \textbf{26.02±0.71} & \textbf{26.30±1.18}           & \textbf{17.19±0.90}           & \textbf{24.05±0.25} & \textbf{16.64±0.41} & \textbf{21.85±0.62}                        & {\color[HTML]{FF0000} \textbf{27.79±2.11}} \\
                                                          & \textbf{ARI}                      & \textbf{13.59±2.02} & \textbf{13.14±1.97} & \textbf{16.34±0.40} & \textbf{24.47±0.13} & \textbf{18.66±1.68}           & \textbf{18.08±0.88}           & \textbf{14.37±0.19} & \textbf{12.21±0.13} & \textbf{19.51±0.45}                        & {\color[HTML]{FF0000} \textbf{26.27±1.96}} \\
\multirow{-4}{*}{\textbf{UAT}}                            & \textbf{F1}                       & \textbf{45.66±1.49} & \textbf{44.22±1.51} & \textbf{46.51±0.17} & \textbf{43.44±0.85} & \textbf{51.32±1.16}           & \textbf{45.19±0.54}           & \textbf{35.72±0.28} & \textbf{35.78±0.38} & \textbf{46.19±0.87}                        & {\color[HTML]{FF0000} \textbf{55.60±1.84}} \\ \hline
\end{tabular}}
\label{compare}
\end{table*}

\subsection{Analysis of Focusing Factor $\beta$}
\begin{figure}[h]
\centering
\small
\begin{minipage}{0.325\linewidth}
\centerline{\includegraphics[width=1\textwidth]{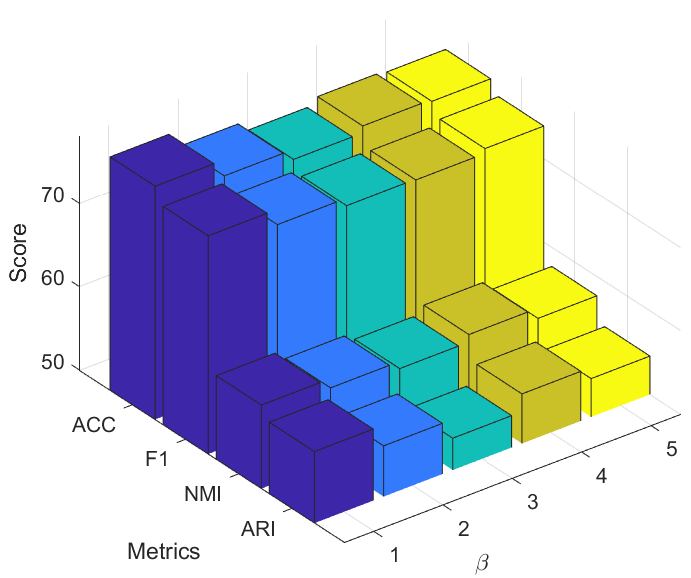}}
\vspace{3pt}
\centerline{(a) CORA}
\centerline{\includegraphics[width=\textwidth]{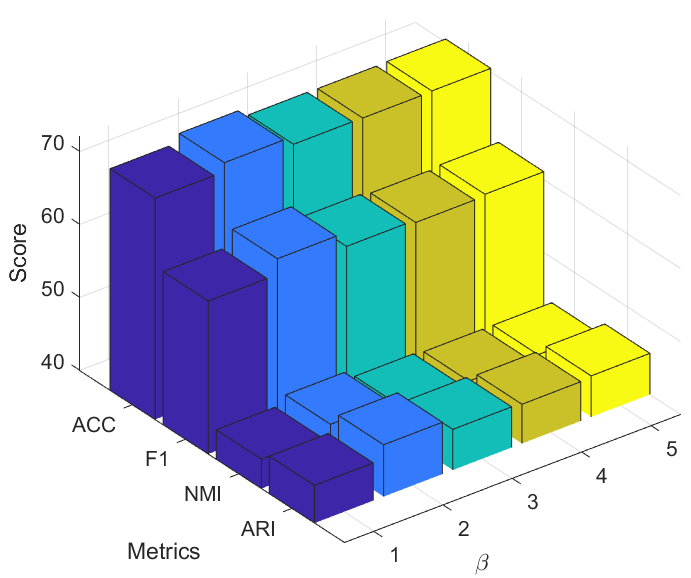}}
\vspace{3pt}
\centerline{(b) CITE}
\end{minipage}
\begin{minipage}{0.325\linewidth}
\centerline{\includegraphics[width=\textwidth]{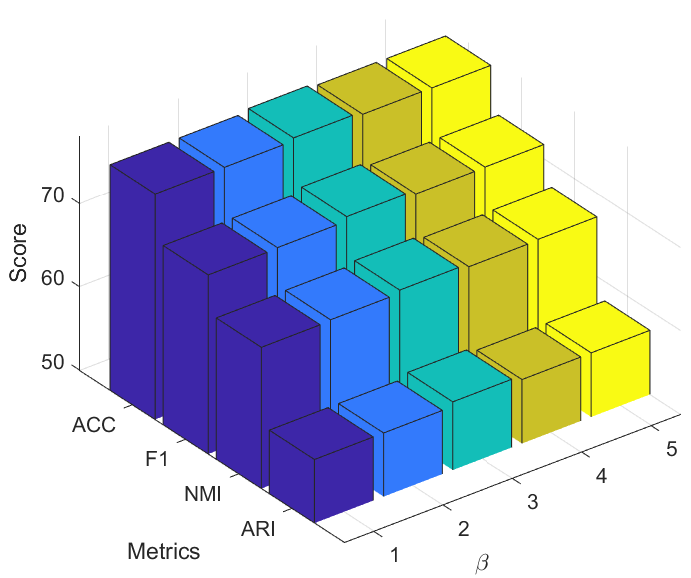}}
\vspace{3pt}
\centerline{(c) AMAP}
\centerline{\includegraphics[width=1\textwidth]{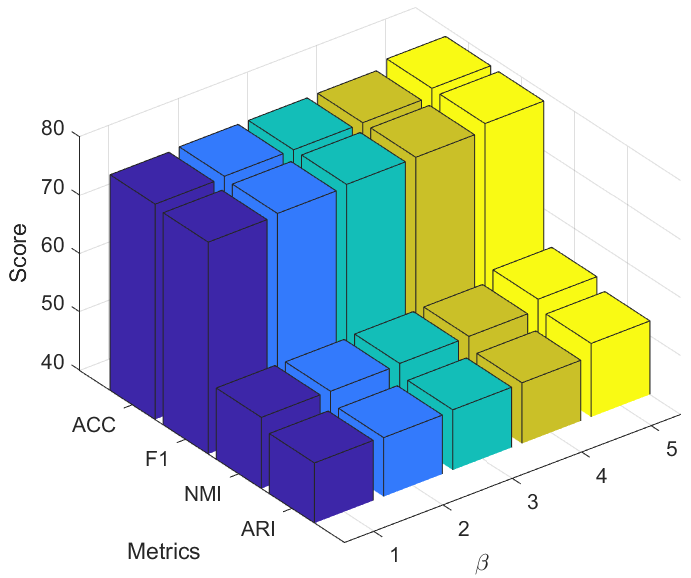}}
\vspace{3pt}
\centerline{(d) BAT}
\end{minipage}
\begin{minipage}{0.325\linewidth}
\centerline{\includegraphics[width=\textwidth]{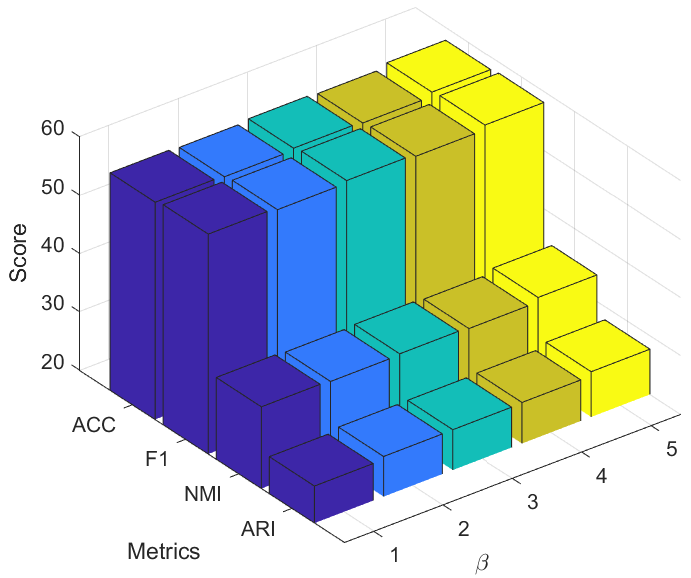}}
\vspace{3pt}
\centerline{(e) EAT}
\vspace{3pt}
\centerline{\includegraphics[width=\textwidth]{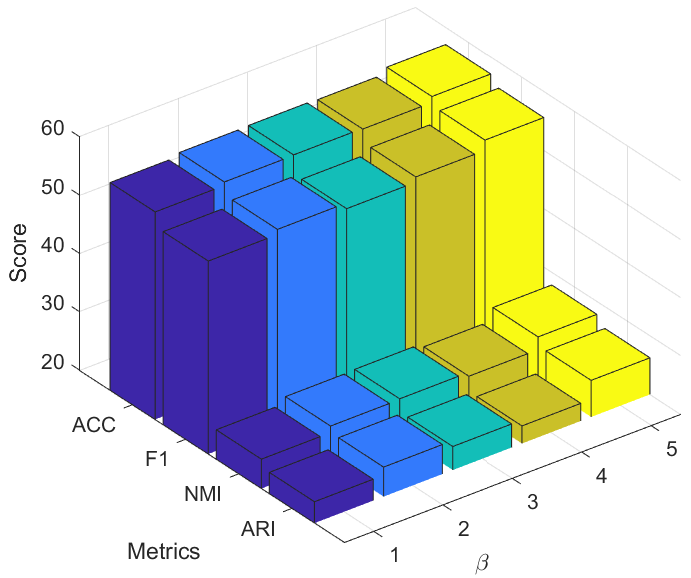}}
\vspace{3pt}
\centerline{(f) UAT}
\end{minipage}
\caption{Sensitivity analysis of the hyper-parameter $\beta$.}
\label{beta}
\end{figure}

\begin{figure}[h]
\centering
\small
\begin{minipage}{0.49\linewidth}
\centerline{Positive Sample Pairs}
\centerline{\includegraphics[width=0.9\textwidth]{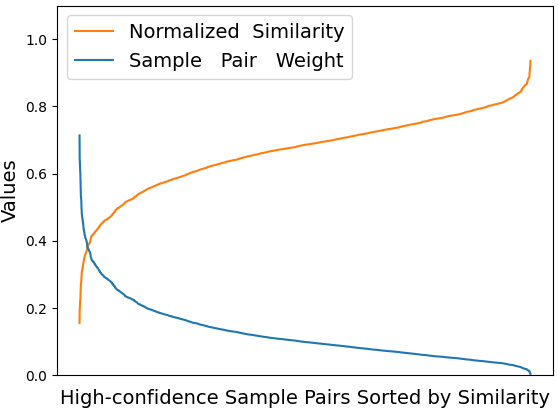}}
\centerline{(a) $\beta=2$}
\vspace{3pt}
\centerline{\includegraphics[width=0.9\textwidth]{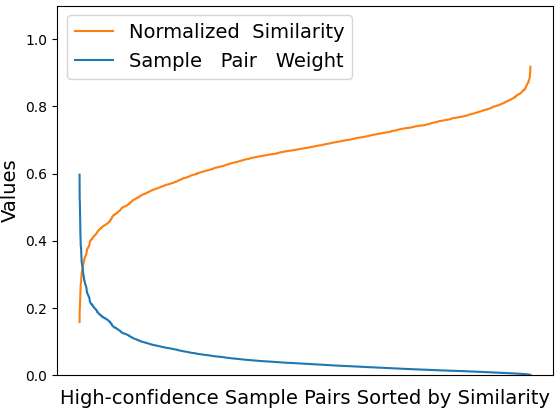}}
\centerline{(b) $\beta=3$}
\vspace{3pt}
\centerline{\includegraphics[width=0.9\textwidth]{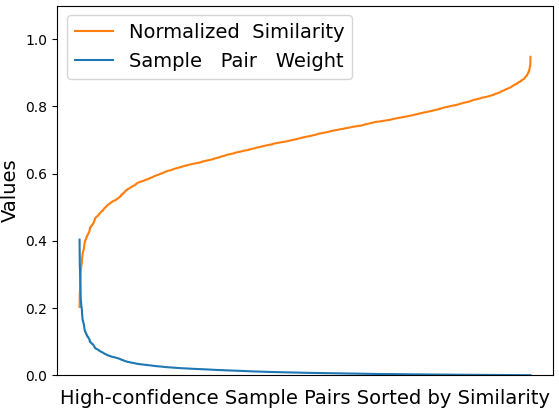}}
\centerline{(c) $\beta=4$}
\vspace{3pt}
\centerline{\includegraphics[width=0.9\textwidth]{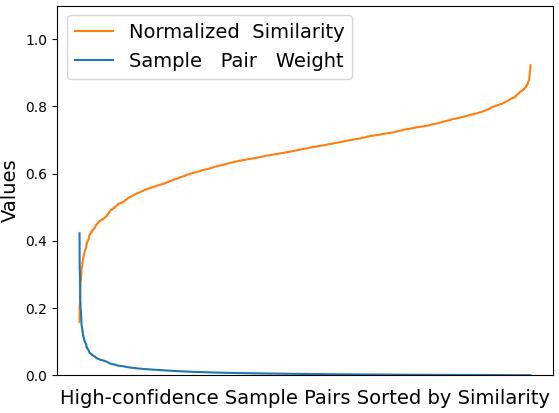}}
\centerline{(d) $\beta=5$}
\end{minipage}
\begin{minipage}{0.49\linewidth}
\centerline{Negative Sample Pairs}
\centerline{\includegraphics[width=0.9\textwidth]{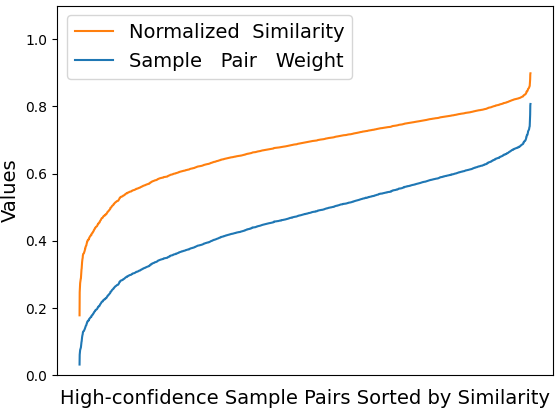}}
\centerline{(e) $\beta=2$}
\vspace{3pt}
\centerline{\includegraphics[width=0.9\textwidth]{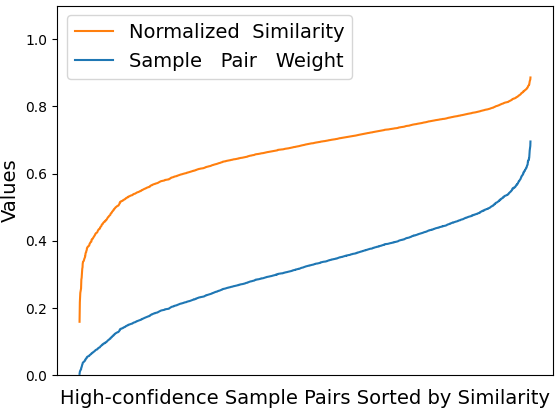}}
\centerline{(f) $\beta=3$}
\vspace{3pt}
\centerline{\includegraphics[width=0.9\textwidth]{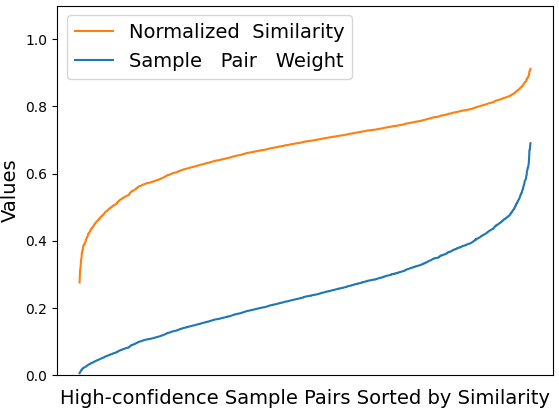}}
\centerline{(g) $\beta=4$}
\vspace{3pt}
\centerline{\includegraphics[width=0.9\textwidth]{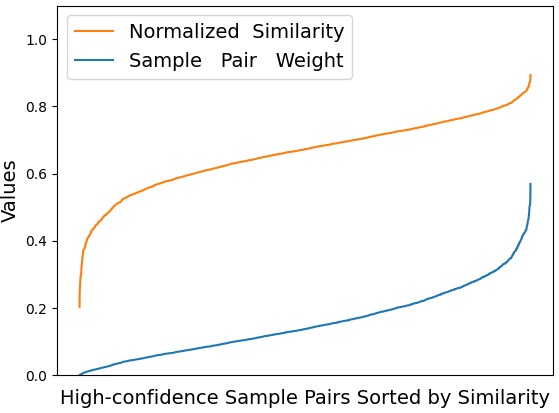}}
\centerline{(h) $\beta=5$}
\end{minipage}
\caption{Visualization of the normalized similarity and the sample pair weight under different value of $\beta$ on CITE dataset.}
\label{beta_cite}
\end{figure}

\begin{figure}[h]
\centering
\small
\begin{minipage}{0.49\linewidth}
\centerline{Positive Sample Pairs}
\centerline{\includegraphics[width=0.9\textwidth]{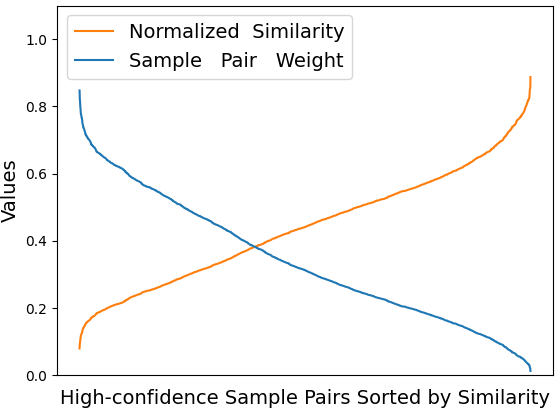}}
\centerline{(a) $\beta=2$}
\vspace{3pt}
\centerline{\includegraphics[width=0.9\textwidth]{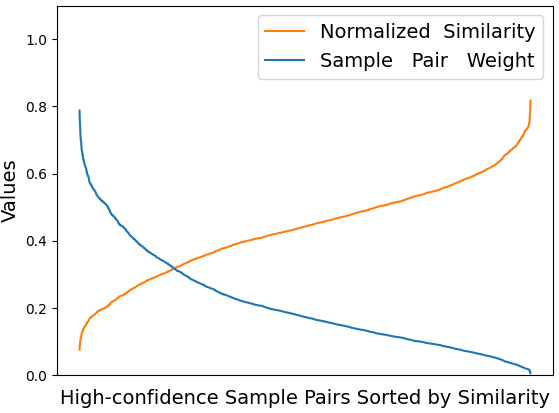}}
\centerline{(b) $\beta=3$}
\vspace{3pt}
\centerline{\includegraphics[width=0.9\textwidth]{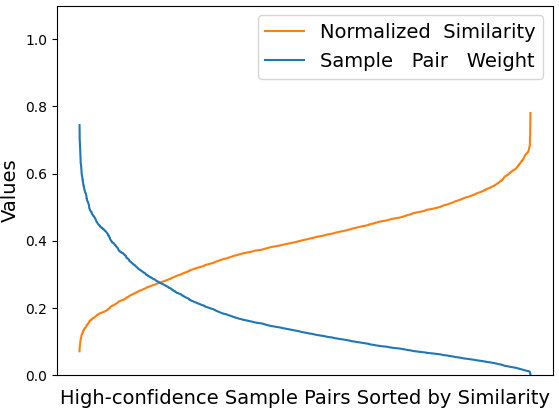}}
\centerline{(c) $\beta=4$}
\vspace{3pt}
\centerline{\includegraphics[width=0.9\textwidth]{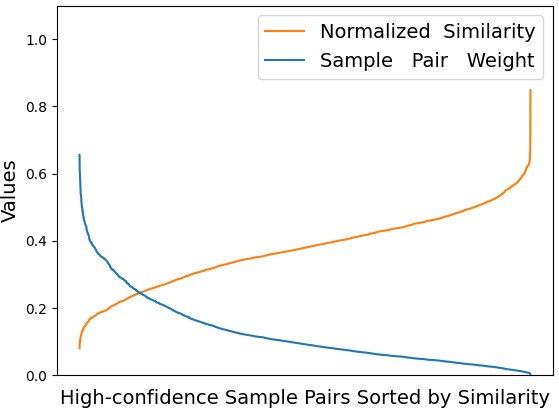}}
\centerline{(d) $\beta=5$}
\end{minipage}
\begin{minipage}{0.49\linewidth}
\centerline{Negative Sample Pairs}
\centerline{\includegraphics[width=0.9\textwidth]{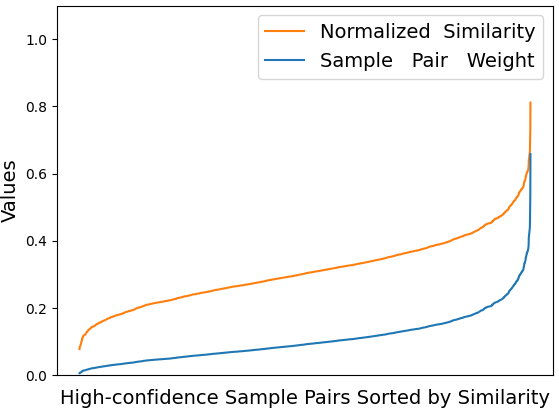}}
\centerline{(e) $\beta=2$}
\vspace{3pt}
\centerline{\includegraphics[width=0.9\textwidth]{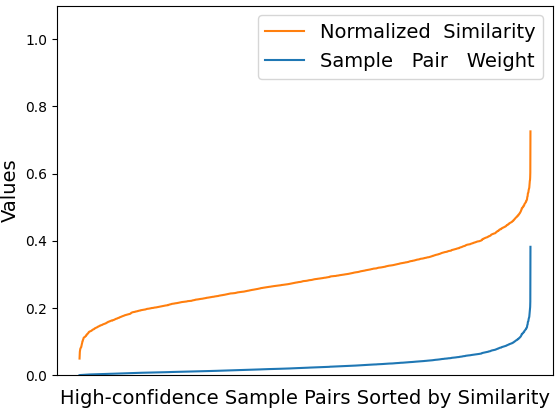}}
\centerline{(f) $\beta=3$}
\vspace{3pt}
\centerline{\includegraphics[width=0.9\textwidth]{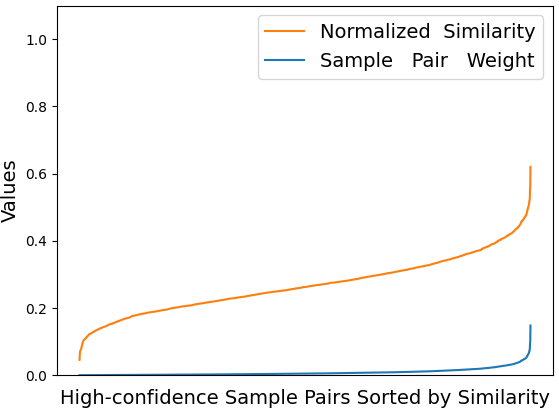}}
\centerline{(g) $\beta=4$}
\vspace{3pt}
\centerline{\includegraphics[width=0.9\textwidth]{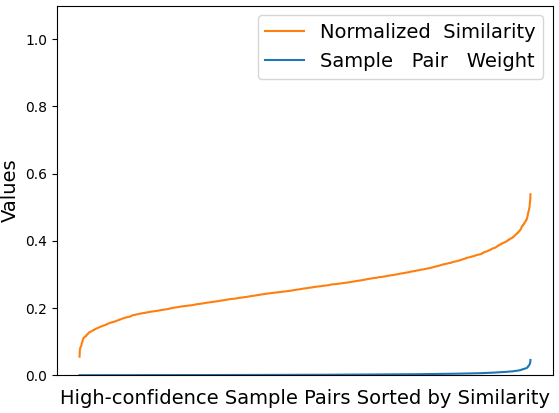}}
\centerline{(h) $\beta=5$}
\end{minipage}
\caption{Visualization of the normalized similarity and the sample pair weight under different value of $\beta$ on CORA dataset.}
\label{beta_cora}
\end{figure}

Besides, the focusing factor $\beta$ in $\mathcal{M}$ is searched in the set $\{1,2,3,4,5\}$ in Fig. \ref{beta}. From these results, we find that our method performs better when $\beta \in [1,2]$. And xxx is insensitive to $\beta$. Besides, when $\beta$ is larger, the down-weighting rate of easy sample pairs will be higher. We verify this property by visualization experiments in Fig. 1 and Fig. 2 of Appendix.
In this section, we conduct visualization experiments to analyze the property of focusing factor hyper-parameter $\beta$ on CITE and CORA datasets. As shown in \ref{beta_cite} and \ref{beta_cora}, to the positive sample pairs, when $\beta$ increases, the down-weighting rate of easy samples also increases. Besides, a similar conclusion can be deduced from the negative sample pairs. 

\subsection{Analysis of Learnable Trade-off Parameter $\alpha$}
\begin{figure}[h]
\centering
\small
\begin{minipage}{0.48\linewidth}
\centerline{\includegraphics[width=1\textwidth]{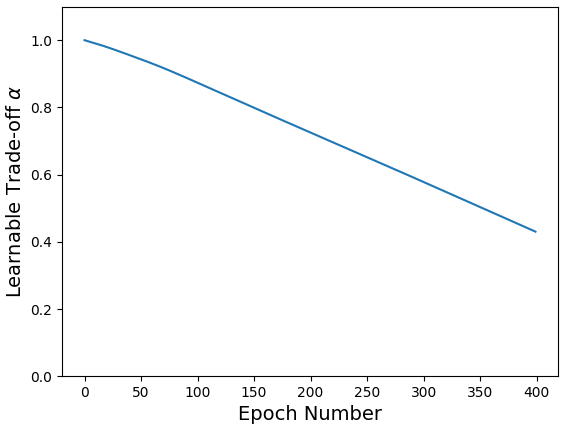}}
\centerline{(a) CORA}
\vspace{3pt}
\end{minipage}
\begin{minipage}{0.48\linewidth}
\centerline{\includegraphics[width=1\textwidth]{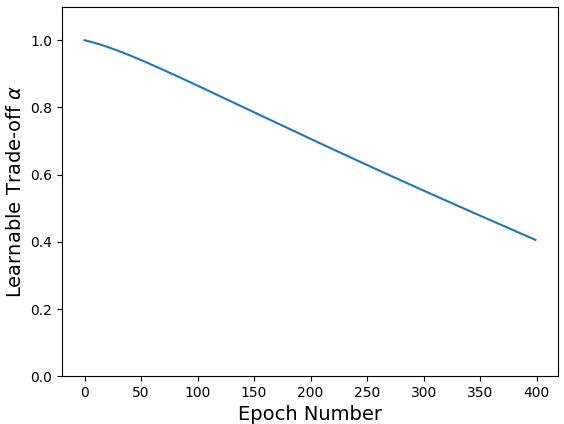}}
\centerline{(b) CITE}
\vspace{3pt}
\end{minipage}
\caption{Learnable trade-off parameter analysis on CORA and CITE datasets.}
\label{alpha}
\end{figure}

Moreover, we analyze the learnable trade-off parameter $\alpha$ in the attribute-structure similarity function $\mathcal{M}$. Concretely, we set $\alpha$ as $0.99999$ as an initialization. As shown in Fig. \ref{alpha}, we observe that $\alpha$ is optimized to around $0.4$ on CORA and CITESEER datasets.

\bibliography{aaai23}